\documentclass[10pt,journal,compsoc]{IEEEtran}
\usepackage{caption}  % 引入 caption 宏包
\usepackage{cite}
\usepackage{graphicx}
\usepackage[lowtilde]{url}
\usepackage[caption=false,font=footnotesize]{subfig}
\usepackage{pgfplots}
\pgfplotsset{compat=newest}
\usetikzlibrary{plotmarks}
\usepackage{grffile}
\usepackage{xcolor}
\usepackage{amsmath}  % 提供 \hat 和 \tilde 命令
\usepackage{amsfonts} % 提供 \mathbb 命令
\usepackage{multirow}
\usepackage{enumitem}
\usepackage[euler]{textgreek}
\usepackage{booktabs} % Allows the use of \toprule, \midrule and \bottomrule in tables for horizontal lines
\usepackage{array}
\usepackage{verbatim}

\usepackage{ragged2e}
\usepackage{diagbox}
\usepackage{makecell}
\usepackage{float}

\usepackage[normalem]{ulem} % JC新增包
\usepackage{soul}
\setul{1.2pt}{1pt}

\usepackage[colorlinks=true, linkcolor=blue, urlcolor=blue]{hyperref}

\usepackage{hyperref}
\hypersetup{colorlinks=true, linkcolor=blue, anchorcolor=blue, citecolor=blue}

\makeatletter
\long\def\@makecaption#1#2{\ifx\@captype\@IEEEtablestring%
\footnotesize\begin{center}{\normalfont\footnotesize #1}\\
{\normalfont\footnotesize\scshape #2}\end{center}%
\@IEEEtablecaptionsepspace
\else
\@IEEEfigurecaptionsepspace
\setbox\@tempboxa\hbox{\normalfont\footnotesize {#1.}~~ #2}%
\ifdim \wd\@tempboxa >\hsize%
\setbox\@tempboxa\hbox{\normalfont\footnotesize {#1.}~~ }%
\parbox[t]{\hsize}{\normalfont\footnotesize \noindent\unhbox\@tempboxa#2}%
\else
\hbox to\hsize{\normalfont\footnotesize\hfil\box\@tempboxa\hfil}\fi\fi}
\makeatother

\newcommand{\RN}[1]{%
  \textup{\uppercase\expandafter{\romannumeral#1}}%
}

\newcolumntype{C}[1]{>{\centering\let\newline\\\arraybackslash\hspace{0pt}}m{#1}}

\renewcommand{\justify}{\leftskip=0pt \rightskip=0pt plus 0cm}

\begin{document}
\title{Perceptual Quality Assessment of Trisoup-Lifting Encoded 3D Point Clouds}

\author{Juncheng~Long,~Honglei~Su,~\IEEEmembership{Member,~IEEE,}~Qi~Liu,~Hui~Yuan,~\IEEEmembership{Senior~Member,~IEEE,}~Wei~Gao,~\IEEEmembership{Senior~Member,~IEEE,}~Jiarun~Song,~\IEEEmembership{Member,~IEEE}~and~Zhou~Wang,~\IEEEmembership{Fellow,~IEEE}% <-this % stops a space

\thanks{ 

This work was supported in part by the National Science Foundation of China under Grant (62222110, 62172259, 62401307 and 62311530104), in part by the High-end Foreign Experts Recruitment Plan of Chinese Ministry of Science and Technology under Grant G2023150003L, in part by the Taishan Scholar Project of Shandong Province (tsqn202103001), in part by the Shandong Provincial Natural Science Foundation, China, under Grants (ZR2022MF275, ZR2022QF076, ZR2022ZD38 and ZR2021MF025), and in part by Funds for Visiting and Training of Teachers in Ordinary Undergraduate Universities in Shandong Province.

Juncheng Long, Honglei Su and Qi Liu are with the College of Electronics and Information, Qingdao University, Qingdao, 266071, China (e-mail: jeasonloong@gmail.com, suhonglei@qdu.edu.cn, sdqi.liu@gmail.com).

Hui Yuan is with the School of Control Science and Engineering, Shandong University, Ji'nan, 250061, China (e-mail: huiyuan@sdu.edu.cn).

Wei Gao is with the School of Electronic and Computer Engineering, Peking University, Shenzhen 518055, China, and also with the Peng Cheng Laboratory, Shenzhen 518066, China (e-mail: gaowei262@pku.edu.cn).

Jiarun Song is with the School of Telecommunications Engineering, Xidian University, Xi’an 710071, China (e-mail: jrsong@xidian.edu.cn).

Zhou Wang is with the Department of Electrical and Computer Engineering, University of Waterloo, Waterloo, ON N2L 3G1, Canada (e-mail: zhou.wang@uwaterloo.ca).

Corresponding author: Honglei Su.
}% <-this % stops a space
}

% \markboth{Submitted to IEEE Transactions on Visualization and Computer Graphics}%
% {Shell \MakeLowercase{\textit{et al.}}: Bare Demo of IEEEtran.cls for Journals}

\markboth{PREPRINT OF A MANUSCRIPT CURRENTLY UNDER REVIEW}%
{Shell \MakeLowercase{\textit{et al.}}: Bare Demo of IEEEtran.cls for Journals}

\IEEEtitleabstractindextext{
\begin{abstract}
\justify
No-reference bitstream-layer point cloud quality assessment (PCQA) can be deployed without full decoding at any network node to achieve real-time quality monitoring. In this work, we develop the first PCQA model dedicated to Trisoup-Lifting encoded 3D point clouds by analyzing bitstreams without full decoding. Specifically, we investigate the relationship among texture bitrate per point (TBPP), texture complexity (TC) and texture quantization parameter (TQP) while geometry encoding is lossless. Subsequently, we estimate TC by utilizing TQP and TBPP. Then, we establish a texture distortion evaluation model based on TC, TBPP and TQP. Ultimately, by integrating this texture distortion model with a geometry attenuation factor, a function of trisoupNodeSizeLog2 (tNSL), we acquire a comprehensive NR bitstream-layer PCQA model named streamPCQ-TL. In addition, this work establishes a database named WPC6.0, the first and largest PCQA database dedicated to Trisoup-Lifting encoding mode, encompassing 400 distorted point clouds with both 4 geometric multiplied by 5 texture distortion levels. Experiment results on M-PCCD, ICIP2020 and the proposed WPC6.0 database suggest that the proposed streamPCQ-TL model exhibits robust and notable performance in contrast to existing advanced PCQA metrics, particularly in terms of computational cost. The dataset and source code will be publicly released at \href{https://github.com/qdushl/Waterloo-Point-Cloud-Database-6.0}{\textit{https://github.com/qdushl/Waterloo-Point-Cloud-Database-6.0}}.
\end{abstract}

}

\maketitle

\IEEEpeerreviewmaketitle

\section{Introduction}\label{sec:introduction}

\IEEEPARstart {T}{hree} dimensional point clouds (3D PCs) are assemblages of points defined in a 3D coordinate system, representing the external surfaces of objects or scenes. Typically harvested via 3D scanning technologies like LiDAR sensors, photogrammetry and depth sensors~\cite{ak2024basics, su2019perceptual}, these points are characterized by their X, Y and Z coordinates and may also encompass additional attributes like color, surface normals and reflectance. The utility of 3D PCs extends to various fields, including autonomous vehicles, digital museums, smart cities, virtual reality, augmented reality, immersive communication and the preservation of cultural heritage~\cite{ak2024basics, su2019perceptual, liu2023point, liu2022perceptual, gu20193d, liu2021reduced, yang2022no}. In these applications, to ensure the quality of user experience, it is essential to conduct real-time monitoring of PC quality for proactively tackling potential contingencies, thereby providing a superior visual experience for users. Given the substantial data volume of 3D PCs, compression is frequently required to store and transmit data efficiently in applications of PCs. Nonetheless, the process of compression can inadvertently introduce a decrease in perceptual quality. Consequently, the development of efficient point cloud quality assessment (PCQA)~\cite{su2019perceptual, liu2023point, liu2022perceptual} method is of significance. It is not only critical for PCQA but also serves as a valuable reference for the design and optimization of PC processing systems.
% streamPCQ-TL系统框图
\begin{figure}[t]
\centering
\includegraphics[width=\linewidth]{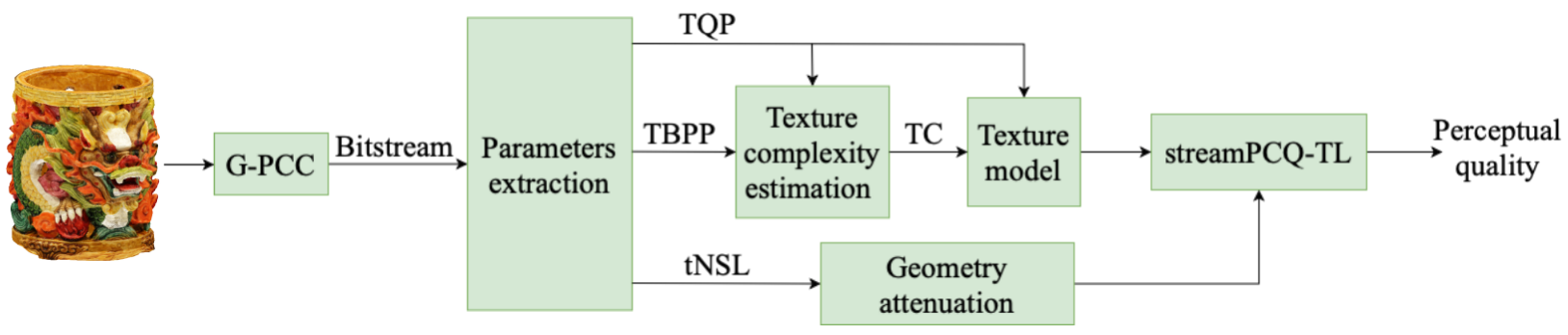}
\caption{Framwork of the proposed model.}
\label{fig:Framework}
\end{figure}

In the year 2017, the Moving Picture Experts Group (MPEG) issued a novel standard~\cite{graziosi2020overview, schwarz2018emerging, cao20193d, chen2023introduction} for 3D graphics coding, encompassing Geometry-based Point Cloud Compression (G-PCC) and Video-based Point Cloud Compression (V-PCC), designed for static and dynamic PCs, respectively. Within the framework of G-PCC, three geometry encoding modes (Octree, Predictive and Trisoup) and three attribute encoding modes (Predicting Transform, Lifting Transform and Region Adaptive Hierarchical Transform) were developed. In this work, we concentrate on the compression of static PCs and consequently select Trisoup-Lifting combination for G-PCC encoding process.

In the domain of objective PCQA, similar to image/video quality assessment, methods are categorized into three distinct paradigms, full-reference (FR), reduced-reference (RR) and no-reference (NR) approaches, based on the accessibility of a pristine reference PC. Within the realm of NR-PCQA, metrics are delineated into media-layer and bitstream-layer models. The media-layer approaches are highly time-consuming due to the necessity of fully decoding PCs, whereas the bitstream-layer approaches operate directly on compressed bitstreams, thereby eliminating the need for fully decoding. Compared to media-layer models, bitstream-layer models possess advantages in terms of low time complexity, making them more suitable for applications requiring stringent real-time performance.

In this paper, we make the first attempt to introduce a PCQA model named streamPCQ-TL dedicated to Trisoup-Lifting encoding mode. The framework of proposed model is illustrated in Fig.~\ref{fig:Framework}. Without fully decoding, the model employs TQP, TBPP and tNSL to estimate coding distortion. And TQP, TBPP and tNSL can be extracted from attribute or geometry bitstream during Trisoup-Lifting decoding. We investigate the relationship among TQP, TBPP and TC, a parameter, characterizes the richness of the texture of PCs, and consequently predict TC based on TQP and TBPP. This forms the logic for developing a texture distortion model. Furthermore, we introduce geometry attenuation factor, a function of tNSL, characterizes geometry distortion. Subsequently, we acquire the integral model streamPCQ-TL by combining the two parts, the texture distortion model and geometry attenuation factor. The major contributions of this work are as follows:
% \hangindent=1em
\begin{enumerate}[leftmargin=2em]
\item We established the first and largest PCQA database dedicated to Trisoup-Lifting encoded 3D PCs, which comprises 400 compressed PCs with their corresponding Mean Opinion Scores (MOSs).%\\
\item We developed a NR PCQA model to predict MOS with three feature parameters (TQP, TBPP and tNSL) extracted from PC bitstreams without fully decoding. To the best of our knowledge, this is the first PCQA model dedicated to Trisoup-Lifting encoded PCs.%\\
\item We demonstrated the effectiveness and robustness of the proposed model and further validate it with other state-of-the-art PCQA metrics on three PCQA databases (WPC6.0, M-PCCD~\cite{alexiou2019comprehensive} and ICIP2020~\cite{perry2020quality}).
\end{enumerate}

The rest of this paper is organized as follows. In Section~\ref{sec:related work}, we discuss the related work. In section~\ref{sec:Subjective Quality Assessment}, we elucidate how to conduct subjective perceptual quality assessment of PCs and establish the WPC6.0 database. In section~\ref{sec:bitstreamPCQA}, we present the technical details of the proposed streamPCQ-TL model. We illustrate experiment results and discussions in section~\ref{sec:Experimental Results and Discussion} and draw conclusions in section~\ref{sec:conclusion}.

\section{Related Work}\label{sec:related work}
In accordance with image/video quality assessment, objective PCQA models can be categorized into FR, RR and NR models contingent upon the availability of reference PC.

\subsection{FR Models}\label{sec:FR-PCQA Models}
Within the realm of FR-PCQA, metrics can be further sorted into three classifications: point-based, projection-based and feature-based. Specifically, point-to-point metrics~\cite{tian2017geometric, javaheri2020generalized, alexiou2018point} give objective scores by calculating distance such as Euclidean distances, Peak-Signal-to-Noise Ratio (PSNR)~\cite{javaheri2020improving}, Hausdorff Distance~\cite{javaheri2020generalized} and Mahalanobis Distance~\cite{javaheri2020mahalanobis}. Point-to-plane metrics~\cite{tian2017geometric} assess the quality of PCs by projected errors of related points along normal direction.  Alexiou \textit{et al}.~\cite{alexiou2018point} introduced a plane-to-plane metric, which can evaluate the quality of PC through the similarity of local surface between reference PC and distorted PC. However, these metrics focus solely on the geometric features of PCs, neglecting the texture information, which is one of the key attributes of PCs.

Projection-based metrics~\cite{torlig2018novel, yang2020predicting, wu2021subjective, wang2023point, he2022tgp, javaheri2022joint, tu2023pseudo} utilize the projected image information of PC to predict quality. In detail, the work presented in~\cite{torlig2018novel} involves the projection of both original reference and distorted PCs onto the six planes that define their bounding boxes. Subsequently, the quality of projected images is assessed by calculating the mean scores derived from advanced image quality metrics. Yang \textit{et al}.~\cite{yang2020predicting} introduced a method that utilizes perspective projection for mapping onto six distinct planes, enabling the extraction of both global and local features from the depth and color images that result from this projection process. Wu \textit{et al}.~\cite{wu2021subjective} put forward two projection-based PCQA models, comprised of a weighted view projection-based model and a patch projection-based model. Wang \textit{et al}.~\cite{wang2023point} introduced a novel projection-based PC saliency map (PQSM) generation method, where depth information is used to better represent the geometric features of PCs. The PQSM generates the final quality score by a saliency-based pooling strategy. He \textit{et al}.~\cite{he2022tgp} proposed a PCQA method (TGP-PCQA) based on geometry and texture projection. Texture and geometry projection maps are obtained from varied perspectives in this method. Javaheri \textit{et al}.~\cite{javaheri2022joint} propounded a  projection-based PCQA model using joint geometry and color information. Tu \textit{et al}.~\cite{tu2023pseudo} proposed a pseudo-reference metric on basis of joint 3D and 2D distortion description, which can produce a joint texture-geometry distribution comprising texture and geometry projection map, which is used to gauge the joint texture-geometry distortion.

Feature-based models assess the quality of PCs through a comparative analysis of the similarity in relevant features between reference and reconstructed PC. Hua \textit{et al}.~\cite{hua2020vqa} proposed a distortion measurement tactic to abstract the geometry and texture characteristics of colour PCs. The extracted information is utilized to construct the feature vector and predict PC quality. Diniz \textit{et al}. proposed four feature descriptors of PCs, Local Luminance Patterns (LLP)~\cite{diniz2020local}, Local Binary Pattern (LBP)~\cite{diniz2020towards}, Perceptual Color Distance Patterns (PCDP)~\cite{diniz2021novel} and BitDance~\cite{diniz2021color}, combining these the statistics of reference and test PCs to predict the perceptual quality of distorted PC. Zhang \textit{et al}.~\cite{zhang2021fqm} proposed a FR method relied on graph signal features and colour features (FQM-GC). Meynet \textit{et al}.~\cite{meynet2020pcqm} introduced PCQM, a FR objective model that is an optimally-weighted linear combination base on geometry and color features for visual quality evaluation of 3D PCs. Inspired by the Structural Similarity (SSIM) metric~\cite{wang2004image} in image quality assessment, PointSSIM~\cite{alexiou2020towards} metric calculates out the quality scores based on the similarity of characteristic values of reference and reconstructive PC. Yang \textit{et al}.~\cite{yang2020inferring} introduced a metric, GraphSIM designed to estimate the visual quality for distorted PCs, considering human vision system is more sensitive to high spatial-frequency components (e.g., contours and edges). Given GraphSIM does not take into account the multiscale characteristics of human visual system, Zhang \textit{et al}.~\cite{zhang2021ms} subsequently proposed a multiscale metirc, Multiscale Graph Similarity (MS-GraphSIM). Hua \textit{et al}.~\cite{hua2022cpc} designed a FR visual quality evaluation metric for coloured point cloud (CPC) based on geometric segmentation and colour transformation (CPC-GSCT), which investigates geometric and color distortion of CPC. Drawing inspiration from classical mechanics, Yang \textit{et al}.~\cite{yang2022mped} developed a metric, multiscale potential energy discrepancy (MPED), which exerts potential energy variation to quantify PC distortion. Xu \textit{et al}.~\cite{xu2021epes} presented a PCQA model, elastic potential energy similarity (EPES), to quantify the influence of distortion on PC visual quality by comparing the elastic potential energy within the springs of reference PC and itscorresponding degraded version. Chetouani \textit{et al.}~\cite{chetouani2021convolutional} and Tliaba \textit{et al.}~\cite{tliba2022point} proposed FR method based on deep learning for predicting visual quality of PCs. Zhang \textit{et al.}~\cite{zhang2023tcdm} developed a metric which can assess quality of PCs by measuring the complexity of converting degraded PC back to its reference PC. Xu \textit{et al.}~\cite{xu2024compressed} proposed a model which can jointly measure geometry and attribute distortion and systematically collect information of variations in the global and local features. Alexiou \textit{et al.}~\cite{alexiou2024pointpca} introduced a range of perceptually relevant descriptors based on principal component analysis (PCA) decomposition, which are applied to texture and geometry data for FR PCQA.

\subsection{RR Models}\label{sec:RR-PCQA Models}
Compared with FR-PCQA metrics, RR-PCQA metrics do not require full information of the reference PCs. Viola \textit{et al}.~\cite{viola2020reduced} proposed a metric to estimate quality of distorted PCs. In this method, they extract a small set of statistical characteristics from reference PC in the geometry, color and normal vector domain to evaluate the visual deformation of PC. Liu \textit{et al}.~\cite{liu2021reduced} introduced a RR model for predicting the quality of V-PCC compressed PCs. The parameters of this model can be estimated by two color features. Zhou \textit{et al}.~\cite{zhou2023reduced} proposed a RR quality metric for PCs, which is founded on Content-oriented saliency Projection (RR-CAP). This method simplifies reference PC and its deformation version into projected saliency maps with a downsampling processing. Liu \textit{et al}.~\cite{liu2022reduced} proposed a RR-PCQA model R-PCQA to quantify the degradation caused by the lossy compression specifically V-PCC, G-PCC and the Audio Video doding Standard (AVS). Su \textit{et al}.~\cite{su2023support} introduced a novel approach for RR PCQA, which is based on a comprehensive candidate feature pool involving compression, geometry, normal, curvature and luminance features and selects major characteristics by a LASSO estimator.

\subsection{NR Models}\label{sec:NR-PCQA Models}
Unlike FR or RR metrics, NR-PCQA metrics~\cite{tu2022v, liu2023once, zhang2023eep, chai2024ms, laazoufi2024point, xiao2024sisim, wang2024rating, tliba2023pcqa, mu2024hallucinated, tliba2024balancing, zhang2024lmm, guo2024wave, lv2024point} solely extract information from distorted PCs to predict perceptual quality of Human Visual System (HVS). Considering the visual masking effect on geometry and texture aspects, Hua \textit{et al}.~\cite{hua2021bqe} proposed a Blind Quality Evaluator for Colored Point Cloud (CPC) based on Visual Perception (BQE-CVP). Concerning HVS is highly sensitive to structure information, Zhou \textit{et al}.~\cite{zhou2024blind} proposed an objective PC quality index with Structure Guided Resampling (SGR) to estimate the perceptual quality of dense 3D PCs. Liu \textit{et al}.~\cite{liu2021pqa} developed a deep learning-based NR-PCQA method, PQA-Net, comprising a multi-view-based joint feature extraction and fusion (MVFEF) module, a distortion type identification (DTI) module and a quality vector prediction (QVP) module. Tao \textit{et al}.~\cite{tao2021point} proposed a Point cloud projection and Multi-scale feature fusion network based Blind Visual Quality Assessment method (denoted as PM-BVQA) for colored PCs. Zhang \textit{et al}.~\cite{zhang2023evaluating} evaluated the PCs from moving camera videos and investigate the method of addressing PCQA tasks by using video quality assessment (VQA) methods. Providing the compelling performance of deep neural network (DNN) on NR quality assessment metric, Yang \textit{et al}.~\cite{yang2022no} presented a NR PCQA model, the image transferred point cloud quality assessment (IT-PCQA), for 3D PCs. Zhang \textit{et al}.~\cite{zhang2022no} proposed a NR PCQA metric for color 3D models represented by both PC and mesh. Fan \textit{et al}.~\cite{fan2022no} proposed a NR quality assessment metric for colored PCs based on captured video sequences. Liu \textit{et al}.~\cite{liu2022progressive} proposed a progressive knowledge transfer based on human visual perception mechanism for perceived quality evaluation of PCs (PKT-PCQA). To leverage the advantages of both PC and projected image modalities, Zhang \textit{et al}.~\cite{zhang2022mm} proposed a NR Multi-Modal Point Cloud Quality Assessment (MM-PCQA) method. Inspired by the hierarchical perception system and considering the intrinsic attributes of PCs, Liu \textit{et al}.~\cite{liu2023point} proposed a NR model ResSCNN based on sparse convolutional neural network (CNN) to predict the perceived quality of PCs. Shan \textit{et al}.~\cite{shan2023gpa} proposed a NR-PCQA model named Graph convolutional PCQA network (GPA-Net). Zhu \textit{et al}.~\cite{zhu20243dta} proposed a two-stage sampling method that can efficiently calculate the PC quality. They designed a twin-attention-based transformer PCQA model (3DTA), which uses the data of the two-stage sampling method as input and directly outputs the predicted quality score. Zhang \textit{et al.}~\cite{zhang2024gms} proposed a NR projection-based Grid Mini-patch Sampling 3D Model Quality Assessment (GMS-3DQA) method to reduce computational resource consumption and enhance evaluation efficiency. Wang \textit{et al}.~\cite{wang2024zoom} introduced a NR-PCQA method (MOD-PCQA) that integrates multiscale characteristics to elevate PC quality perception across diverse scales and granularities. Liang \textit{et al.}~\cite{liang2023mfe} proposed a NR model based on deep learning, which is consisting of adaptive feature extraction (AFE) module, local quality acquisition (LQA) module and global quality acquisition (GQA) module. Chen \textit{et al}.~\cite{chen2024dynamic} proposed a NR-PCQA method with hypergraph learning. Only considering plain visual and geometric factors, Chai \textit{et al}.~\cite{chai2024plain} proposed an end-to-end learning model named Plain-PCQA for a quantitatively assessing objective method of 3D dense PCs corresponding to human perception. Shan \textit{et al}.~\cite{shan2024contrastive} proposed a pre-training method tailored for NR-PCQA (CoPA), which makes the pre-trained model learn quality-aware representations from unlabeled data. \\
%\newline
\indent Beyond the aforementioned NR-PCQA metrics, there have recently emerged novel bitstream-based NR metrics within the field. Su \textit{et al}.~\cite{su2023bitstream} introduced a bitstream-based NR model for perceived quality assessment of Trisoup-RAHT encoded PCs. Liu \textit{et al}.~\cite{liu2022no} proposed a bitstream-layer NR model for the perceptual quality assessment of V-PCC encoded PCs. Lv \textit{et al.}~\cite{lv2024no} proposed a bitstream-based NR model to evaluate quality of Octreee-Lifting encoded PCs. To a certain extent, these studies address the issue of real-time monitoring of PC quality in network nodes. However, there has been no discussion on the quality assessment of Trisoup-Lifting encoded PCs yet.

\subsection{Summary}\label{sec:Summary}
In conclusion, these PCQA metrics are tailored to specific application contexts. The utilization of FR methods is fully dependent on complete reference PCs while assessing their corresponding distorted PCs. Conversely, NR metrics depend exclusively on the distorted PC for quality evaluation, operating without the need to access reference PC. And RR models use a portion of feature information derived from the reference PCs as reference, rather than relying entirely on the full reference data. Worth noting is the fact that bitstream-based NR metrics, capable of directly extracting feature information from the bitstream for assessing the perceived PC quality, run at the bitstream-layer rather than the media-layer and can obviate the need for fully decoding of compressed PCs, which implies they own the nature low computational consumption, making them a better choice for time-critical applications. However, there is currently no PCQA model specifically designed for Trisoup-Lifting encoded PCs. To fill this gap, we propose this work. For a detailed description, please refer to Section~\ref{sec:Subjective Quality Assessment} and ~\ref{sec:bitstreamPCQA}.

\section{The WPC6.0 Database}
\label{sec:Subjective Quality Assessment}
\subsection{Generation of Distorted PCs}
In order to develop a robust PCQA model tailored to PCs encoded by Trisoup-Lifting mode, capable of tackling broader spectrum of distortion levels, we made the WPC6.0 database rather than using existing ones containing Trisoup-Lifting distorted PCs such as M-PCCD~\cite{alexiou2019comprehensive} and ICIP2020~\cite{perry2020quality}. The latter two, as comprehensive PCQA databases, have a small number of Trisoup-Lifting distorted PCs and do not have a wide range of distortion levels in both geometry and attribute aspects. Therefore, it is not enough rich for them at both distortion levels and the number of distorted PCs. Based on the current situation, we make a decision to create the WPC6.0 database which is the first one dedicated to the PCQA issue for Trisoup-Lifting encoding mode.

We selected~\textit{bag},~\textit{banana},~\textit{biscuits},~\textit{cake},~\textit{cauliflower},~\textit{flowerpot},~\textit{glasses\_case},~\textit{honeydew\_melon},~\textit{house},~\textit{litchi},~\textit{mushroom},~\textit{pen\_container},~\textit{pineapple},~\textit{ping-pong\_bat},~\textit{puer\_tea},~\textit{pumpkin},~\textit{ship},~\textit{statue},~\textit{stone} and ~\textit{tool\_box} from the Waterloo Point Cloud (WPC) database~\cite{liu2022perceptual} for subjective quality evaluation. To encode these PCs, we selected the MPEG G-PCC platform (TMC13 release-v23.0-rc2). For each PC, we made 20 distinct compressed versions by setting four different values (3, 4, 5 and 6) for geometry parameter (i.e., tNSL) and five different values (28, 34, 40, 46 and 51) for texture quantization parameter (i.e., TQP). Other encoding configuration parameters are set to default values. Subsequently, we generated a total of \(20\times4\times5\) = 400 encoded PCs for subjective quality assessment. It should be noted that the WPC6.0 database is the first and largest PCQA database dedicated to Trisoup-Lifting encoding mode possessing a more extensive spectrum of distortion levels.

% 主观实验设置
\begin{table}[t]
\renewcommand{\arraystretch}{1.2}
\centering
% \caption{Rendering parameter settings and display screen parameters}
\caption{Rendering parameter settings.}
\label{tab:subjective evaluation settings}
\scalebox{1.2}{
    \begin{tabular}{c | c }
    \toprule
    \hline
    Setting items & Parameter\\
    \hline
    Render window & \(960\times960\) \\
    Point type & point \\
    % Point size ($\operatorname{PQS}_{0.125, 0.25, 0.5, 1}$) & 8, 4, 2, 1 \\
    Point size & 1 \\
    Virtual camera radius & 5000 \\
    Degree/Point of view & 2 \\
    Frame image & 180 fps \(\times\) 2 \\
    Display time & 10 seconds \\
    The sequence of video playback & random \\
    Screen resolution & \(1920\times1080\) \\
    Screen size & 23.8 inches \\
    Screen refresh rate & 60Hz \\
    \hline
    \bottomrule
    \end{tabular}
}
\end{table}

\subsection{Subjective Quality Evaluation}
Prior to conducting the subjective evaluation, we recruited 30 participants and all subjects had either normal or corrected-to-normal vision. The Technicolor renderer~\cite{guede2017technicolor} was utilized to convert each PC to a video sequence. The rendering settings of the videos and the display specifications utilized in the subjective evaluation are detailed in Table~\ref{tab:subjective evaluation settings}. According to ITU-R Recommendation BT.500-13~\cite{series2012methodology}, the monitor utilized in the subjective evaluation was calibrated. In the subjective evaluation phase, we employed the Double-Stimulus Impairment Scale (DSIS) methodology~\cite{series2012methodology}, wherein both the reference and distorted videos derived from PCs were presented concurrently to observers in a side-by-side comparison. The observers viewed these videos at a distance equivalent to twice the vertical screen dimension and rated the content by a bespoke interface post-playback. Before commencing formal evaluation, a training section was conducted, during which subjects were presented with 12 videos that were distinct from those in the formal evaluation and ensured no content overlap. Therefore, participants were acquainted with the types and degrees of distortion featured in the formal evaluation, thereby limiting the impact of any potential learning effects. Significantly, to enhance the granularity of the rating scale and capture subtler differentiations, we employed a continuous 100-point scale, surpassing the conventional five-category scale. 

% Pic_Subject_MOS_Std
\begin{figure}[t]
\centering
\subfloat[]
{\includegraphics[width=0.49\linewidth]
{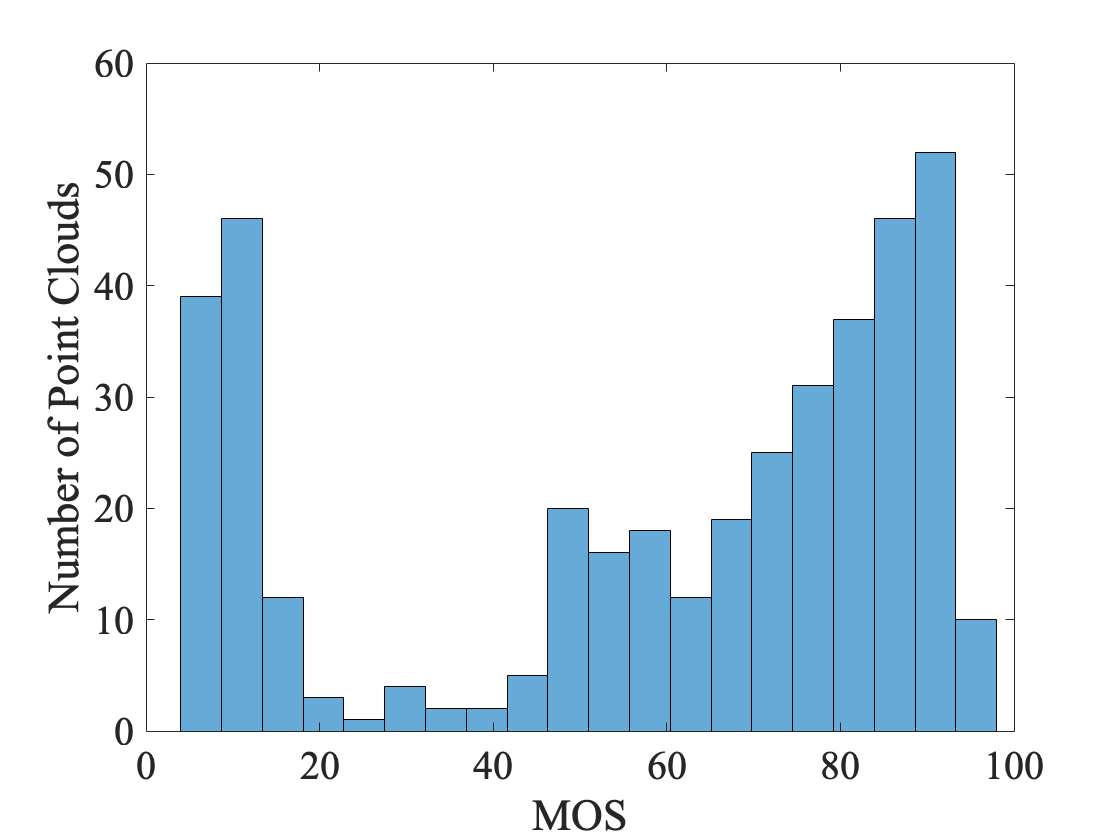}}\hskip0.01em
% \\
\subfloat[]
{\includegraphics[width=0.49\linewidth]
{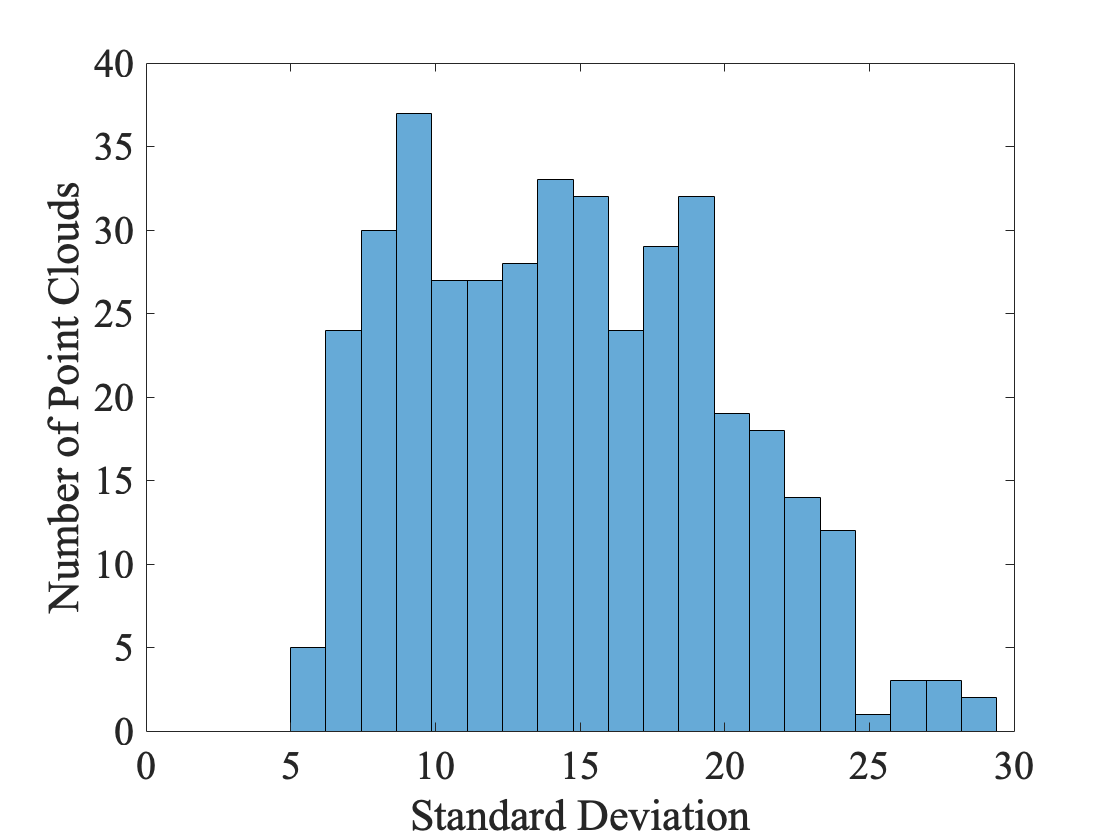}}
\caption{MOS distribution of the WPC6.0 database.}
\label{fig:Subject_MOS_Std}
\end{figure}
% \vspace{0.01em}
% Pic_Subject_PLCC_SRCC
\begin{figure}[t]
\centering
\subfloat[]
{\includegraphics[width=0.49\linewidth]{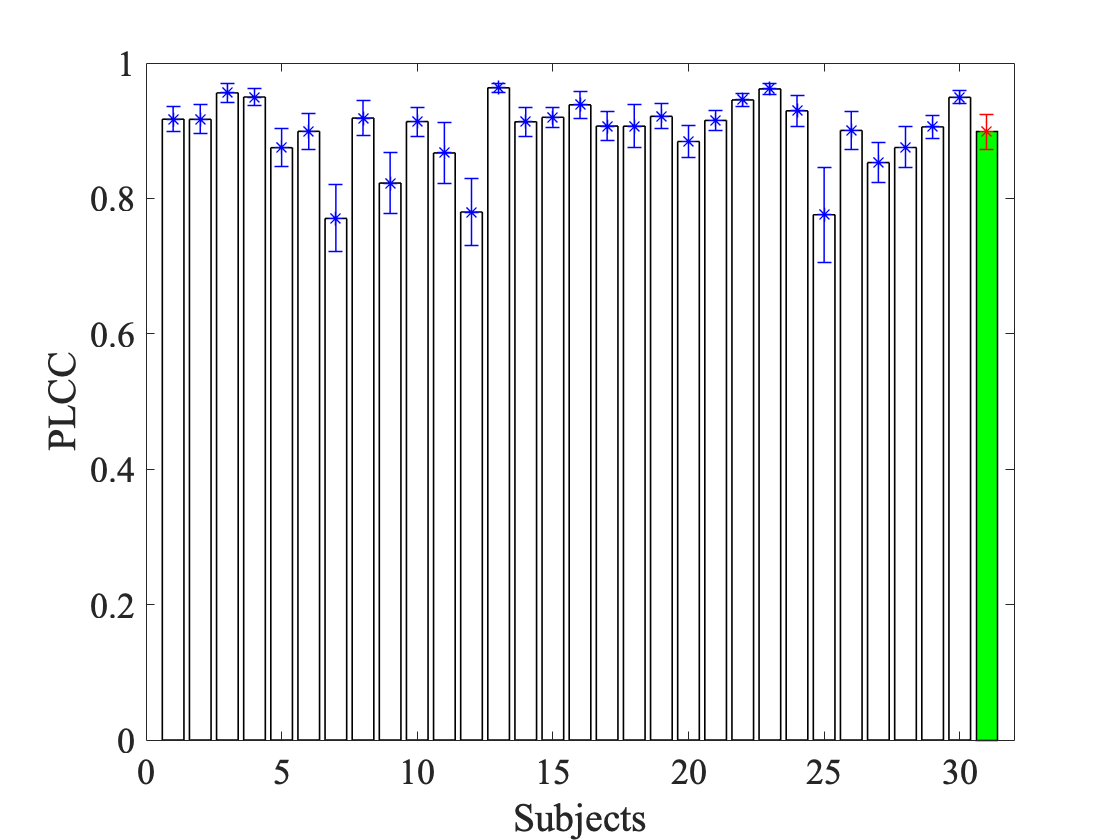}}\hskip0.01em
% \\
\subfloat[]
{\includegraphics[width=0.49\linewidth]{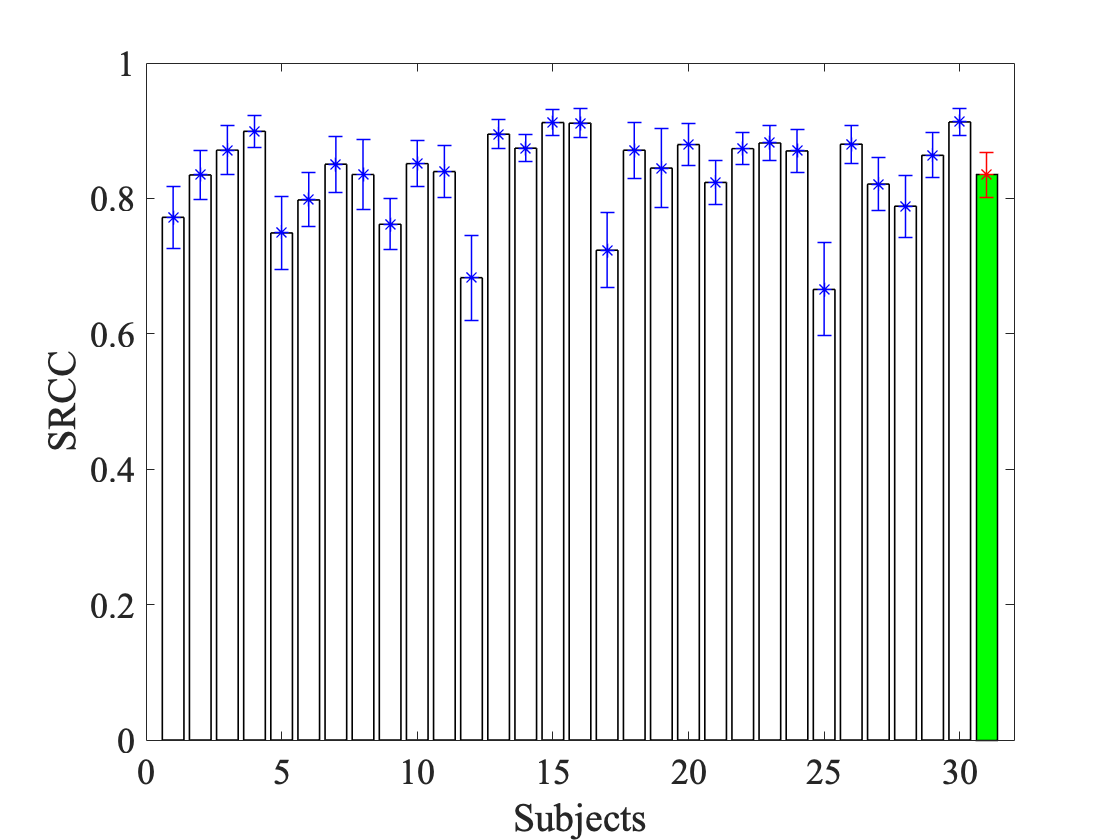}}
\caption{PLCC and SRCC between each subject's ratings and MOSs.}
% Rightmost column: performance of all subjects.
\label{fig:Subject_PLCC_SRCC}
\end{figure}

\subsection{Data Post-Processing and Analysis}
Concerning the subjective scores obtained from participants, to transform data across various scales to a common scale and to diminish biases, we subjected these scores to Z-score standardization. The Z-score~\cite{van1995quality} of the i-th PC rated by the j-th observer is denoted as $Z_{i,j}$ determined by the following formula 
\begin{equation}
    Z_{i,j}=\frac{X_{i,j}-\mu_{x_i}}{\sigma_{x_i}}\ ,
\end{equation}
where $X_{i,j}$ represents the raw score given by the j-th observer to the i-th PC, $\mu_{x_i}$ is the mean of the raw scores for the i-th PC from all participants and $\sigma_{x_i}$ is the standard deviation of the raw scores for the i-th PC among all subjects, respectively. Furthermore, we removed outliers employing the technique proposed in~\cite{series2012methodology}. Subsequently, we obtained Z-scores spanning from 1 to 100. Afterwards, we computed the mean of the Z-scores for each distorted PC deriving the MOS. Fig.~\ref{fig:Subject_MOS_Std} presents the histograms of the MOSs and their associated standard deviations. Due to a drastic degradation in the quality of PCs arising in between tNSL=5 and tNSL=6, the MOSs showed in Fig.~\ref{fig:Subject_MOS_Std}(a) align with the perceptual quality distribution of HVS rather than evenly distributing in 1-100 scale, which correspondingly causes these standard deviations, depicted in Fig.~\ref{fig:Subject_MOS_Std}(b), tending toward large values. Regarding the MOS as the ``ground truth'', the performance of individual subjects is assessed through the computation of the correlation coefficient between each subject's rating (i.e., Z-score) and the MOS for the respective distorted PCs. We selected the Pearson Linear Correlation Coefficient (PLCC) and Spearman Rank-order Correlation Coefficient (SRCC) as criteria for evaluation. Fig.~\ref{fig:Subject_PLCC_SRCC} illustrates the results of PLCC and SRCC. The majority of subjects exhibits consistent behavior in the results. Additionally, the average performance of all subjects is depicted in the rightmost columns of Fig.~\ref{fig:Subject_PLCC_SRCC}.

% MOS_TQP图
% [htpb]
\begin{figure}[t]
\centering
\includegraphics[width=1\linewidth]{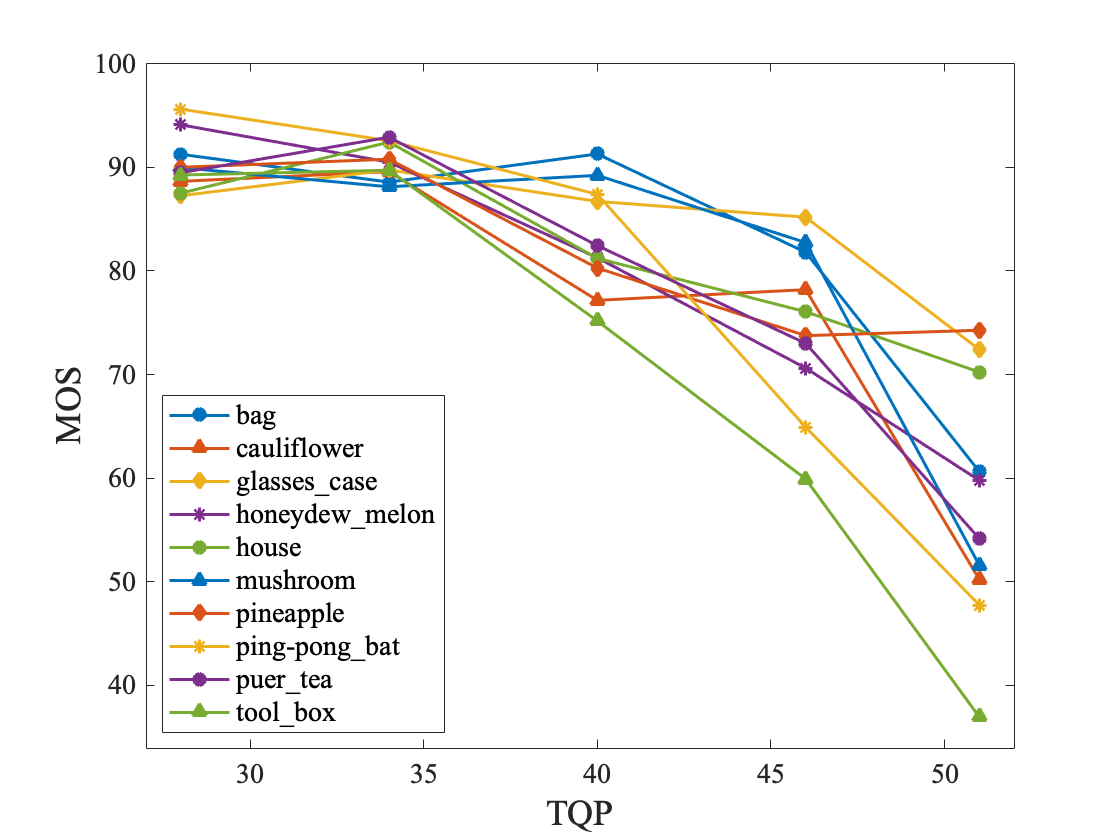}
\caption{Relationship between MOS and TQP.}
\label{fig:MOS_TQP_10}
\end{figure}

\section{Proposed streamPCQ-TL Model}
\label{sec:bitstreamPCQA}
\subsection{Perceptual Texture Distortion and Texture Quantization Parameter}
Our study establishes a texture distortion model in the condition that geometry remains lossless. Consistent with what happens in image and video compression, attribute (texture) lossy compression will incidentally introduce distortions during quantization, leading to the appearance of fuzzy distortions and other unintended compression artifacts. Accordingly, the texture distortion is closely correlated with the quantization step and the corresponding TQP. To investigate the intrinsic relationship, we selected 10 PCs, specifically~\textit{bag},~\textit{cauliflower},~\textit{glasses\_case},~\textit{honeydew\_melon},~\textit{house},~\textit{mushroom},~\textit{pineapple},~\textit{ping-pong\_bat},~\textit{puer\_tea} and~\textit{tool\_box} as training set, compressed by Trisoup-Lifting with constant TQPs. Fig.~\ref{fig:MOS_TQP_10} illustrates an approximate linear relationship between MOS and TQP in the selected PCs. And we also can discover that these curves are varied for distinct PCs. Specifically, it can be noted that PCs possessing more diverse texture attributes such as~\textit{glasses\_case} tend to exhibit a markedly higher MOS at specific TQP levels. Conversely, PCs with less texture information like~\textit{tool\_box} demonstrate a lower MOS under the identical conditions. Considering the texture masking effect of the HVS, this observed discrepancy can be elucidated and implies that the effect is shaped by the level of texture richness. Alternatively speaking, the effect can be described in terms of texture complexity (TC). Consequently, we introduced TC into the texture distortion model.

% TC_TBPP  TQP:28-51
\begin{figure}[t]
\centering
%  \captionsetup{justification=centering}
\subfloat[TQP=28]
{\includegraphics[width=0.49\linewidth]
{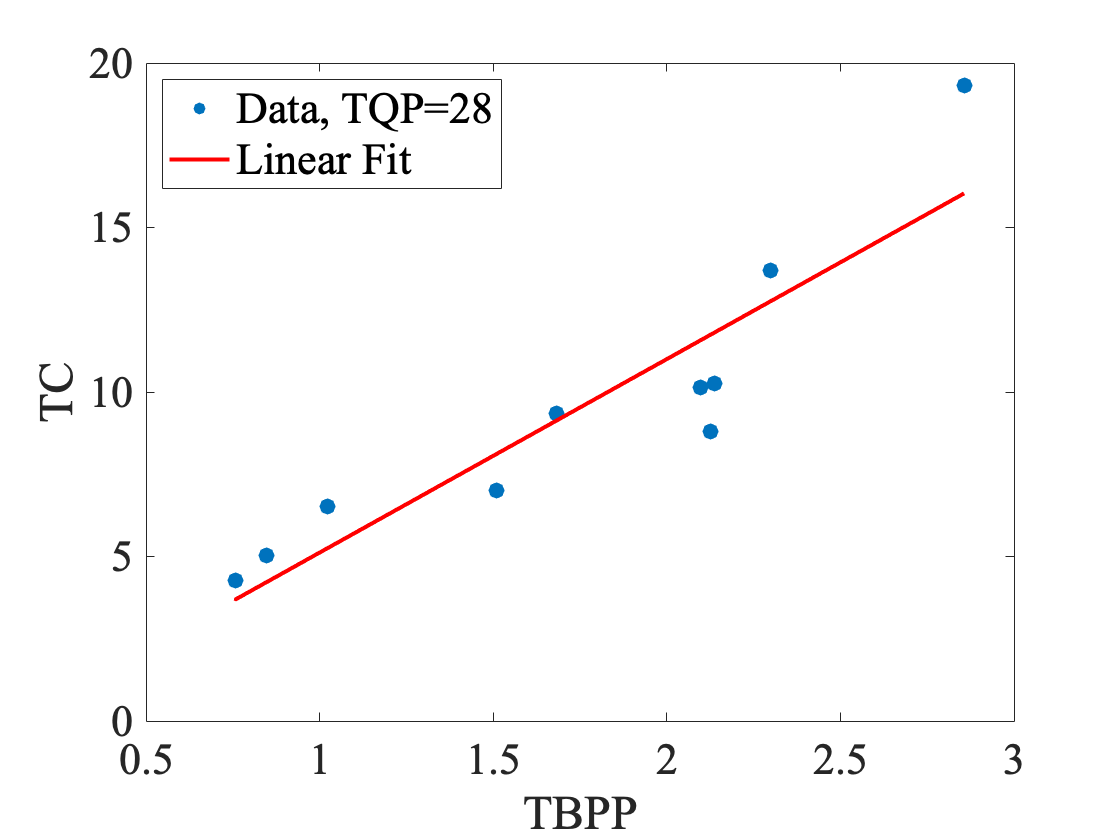}}\hskip0.1em
\subfloat[TQP=34]
{\includegraphics[width=0.49\linewidth]
{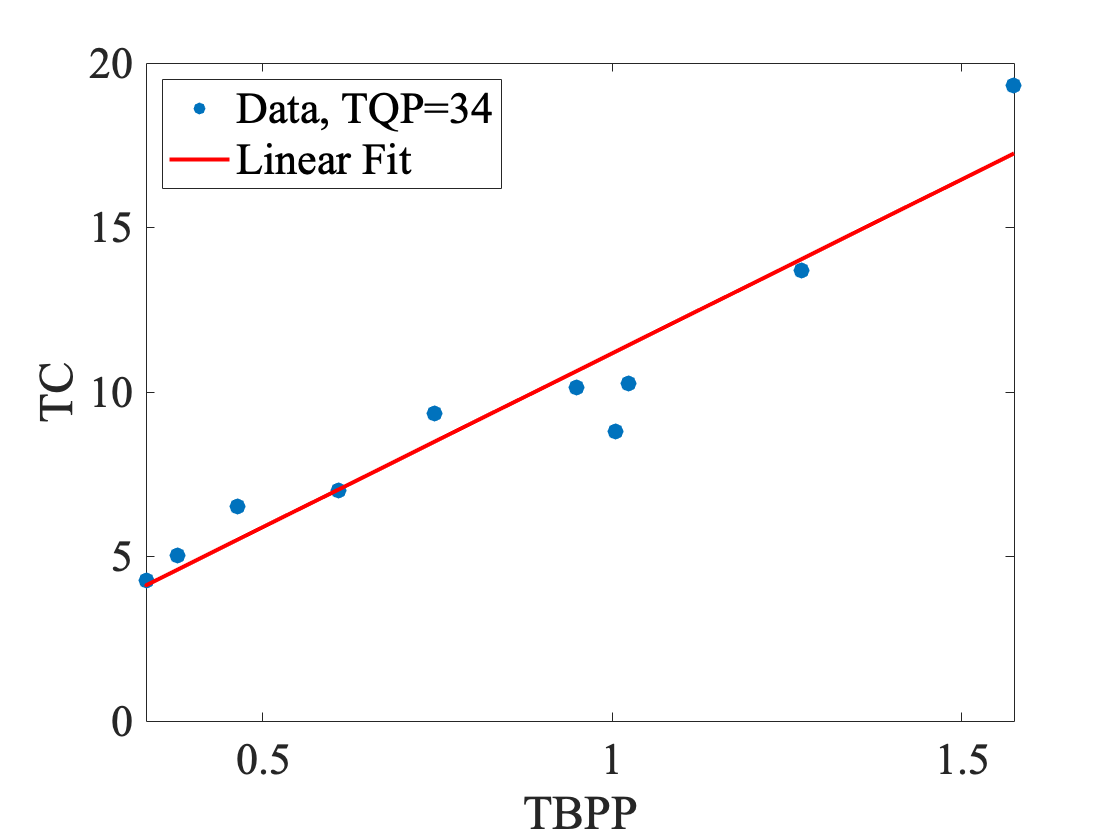}}\hskip0.1em
\subfloat[TQP=40]
{\includegraphics[width=0.49\linewidth]
{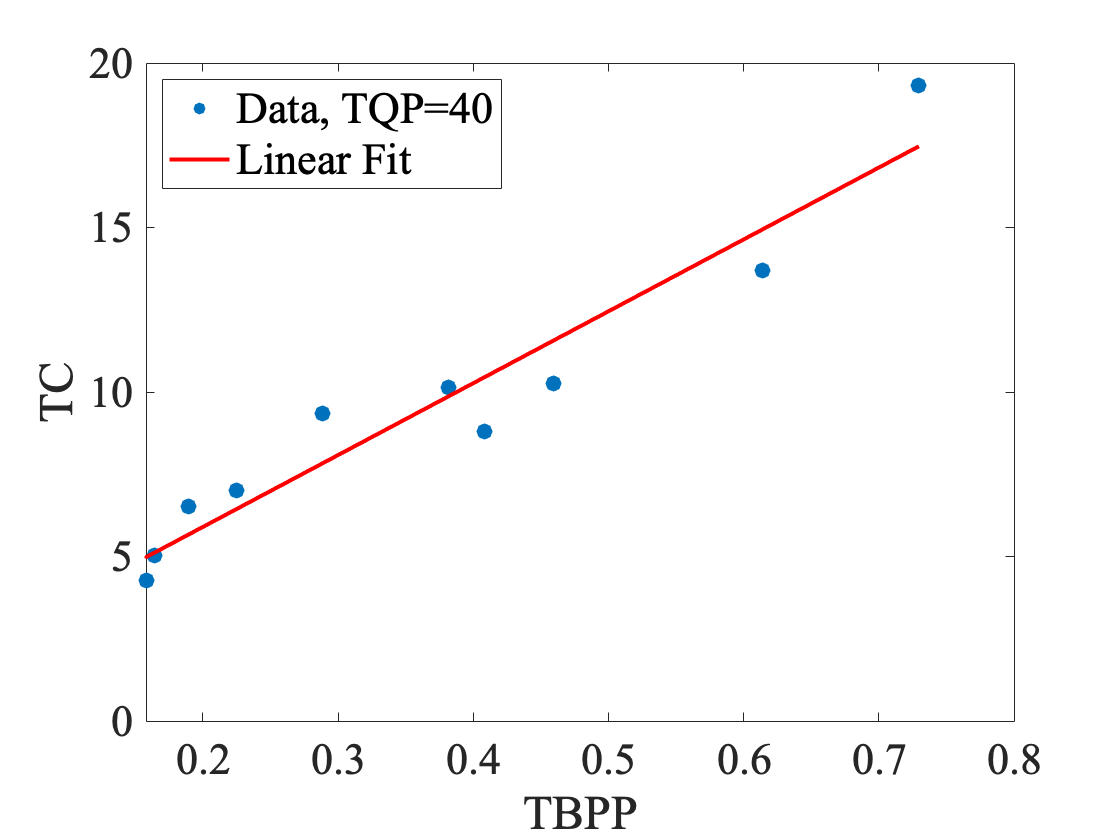}}\hskip0.1em
\subfloat[TQP=46]
{\includegraphics[width=0.49\linewidth]
{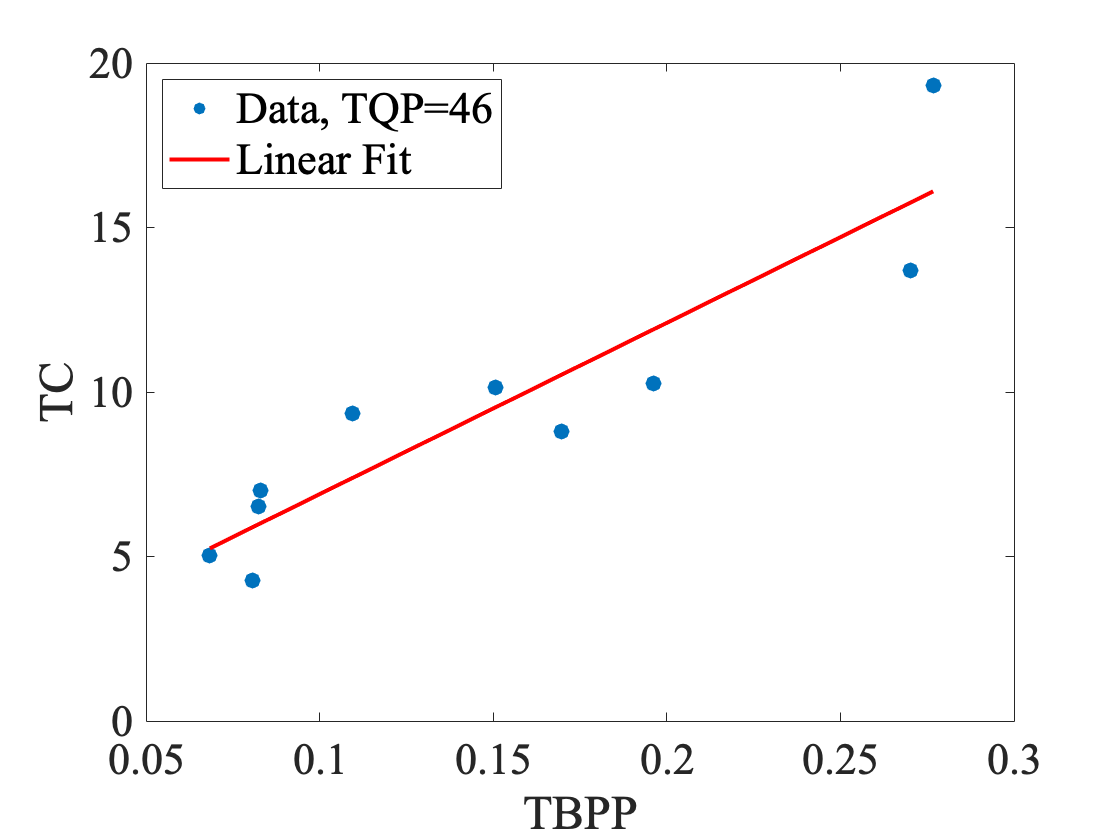}}\hskip0.1em
\subfloat[TQP=51]
{\includegraphics[width=0.49\linewidth]
{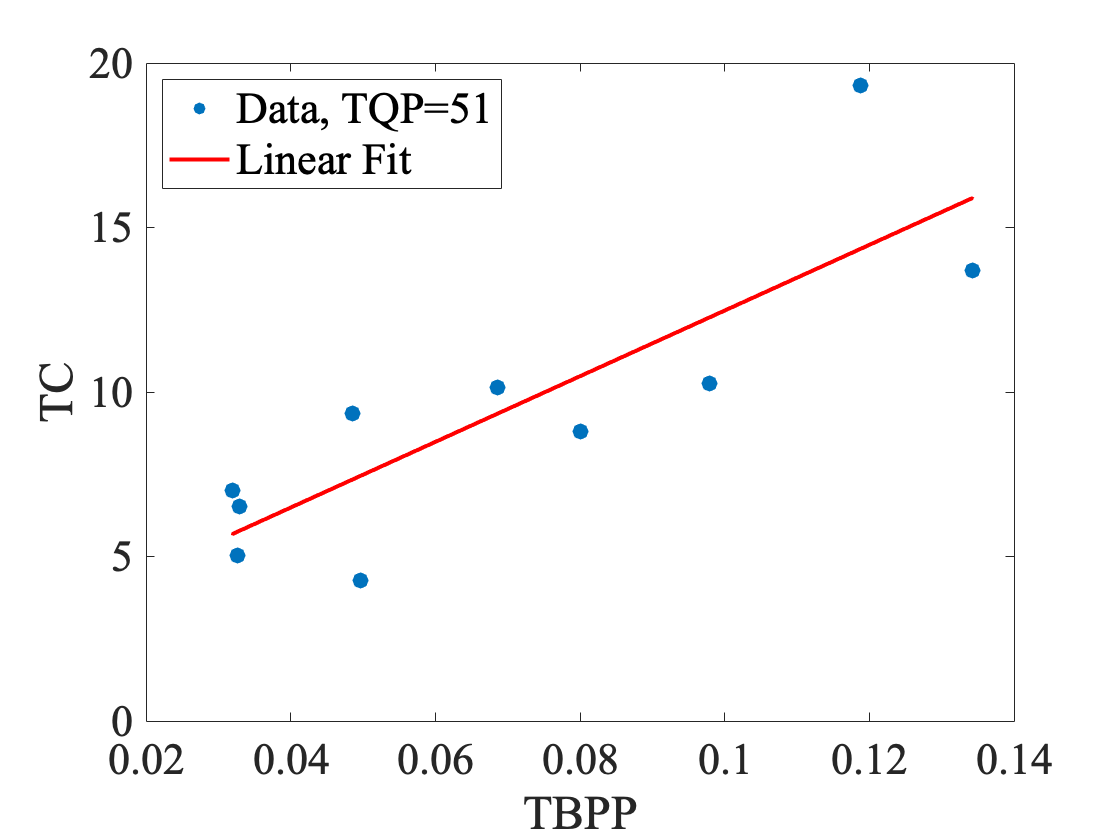}}\hskip0.1em
\caption{TC vs. TBPP corresponding to different TQPs.}
\label{fig:TC_TBPP}
\end{figure}

\begin{table}[t] 
\renewcommand{\arraystretch}{1.3}
\centering
\caption{PLCC between TC and TBPP.} 
\label{tab:TC_TBPP_PLCC}
\begin{tabular}{c | c c c c c}  
\toprule
\hline
TQP & 28 & 34 & 40 & 46 & 51\\
\hline  
PLCC  & 0.9658 & 0.9780 & 0.9781 & 0.9623 & 0.9305\\
\hline
\bottomrule
\end{tabular}
\end{table}

% [htpb]
% TCTBPP_Fit_TQP
\begin{figure}[t]
\centering
\subfloat[]
{\includegraphics[width=0.49\linewidth]
{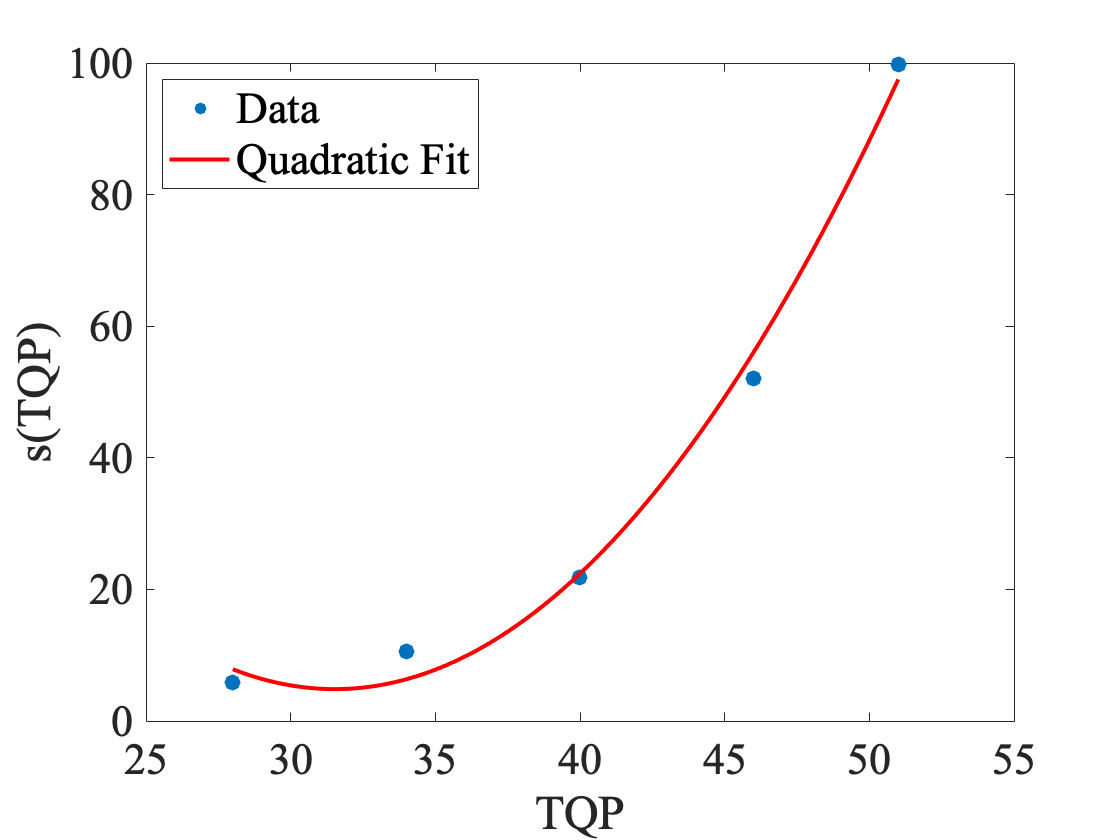}}\hskip0.01em
% \\
\subfloat[]
{\includegraphics[width=0.49\linewidth]
{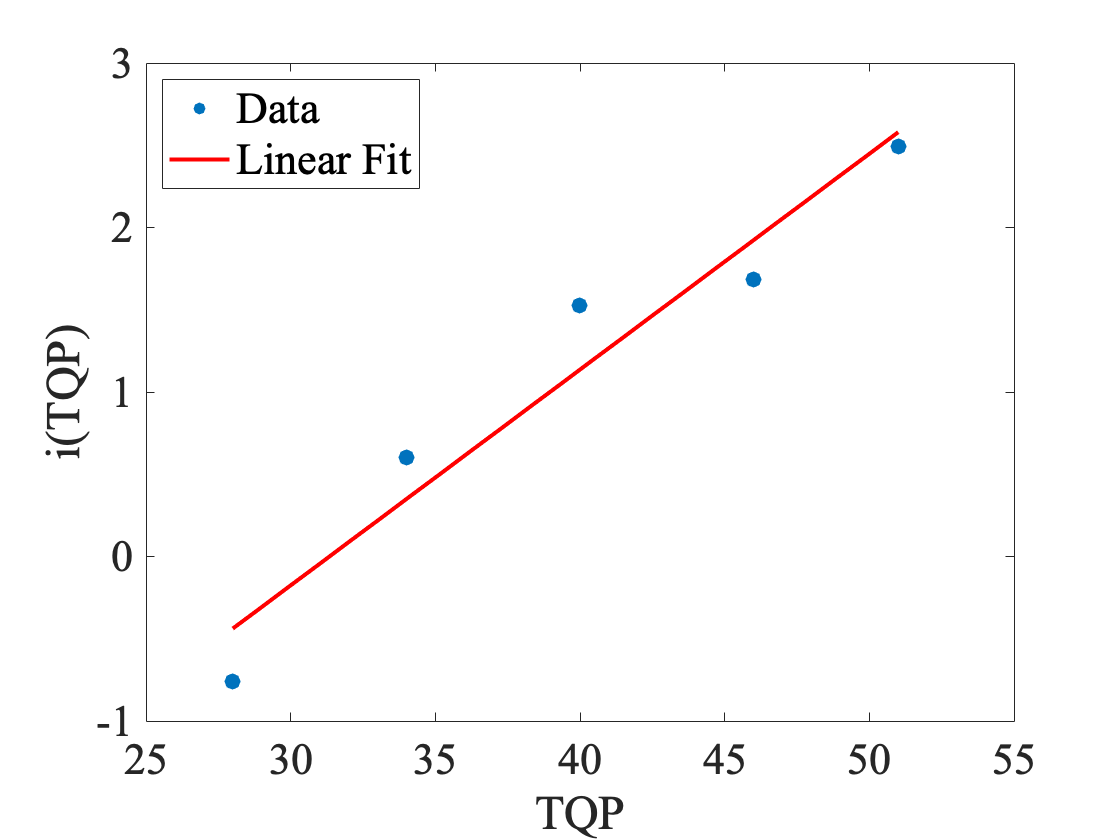}}
\caption{Slope and intercept for various TQPs.}
\label{fig:TCTBPP_Fit_TQP}
\end{figure}

\subsection{Texture Complexity Estimation}
In instances where attribute values from reference PC are accessible, TC can be derived by calculating the mean standard deviation of some values within localized blocks, a method previously detailed in~\cite{clarke1985transform}. Taking into account that the absence of available reference PC in NR-PCQA, we estimate TC using parameters, TBPP and TQP, extracted from bitstreams of compressed PCs. Therefore, with a constant  TQP, a linear relationship between TC and TBPP can be approximated by the following equation: 
\begin{equation}
    TC=s(TQP)\cdot TBPP+i(TQP)\ ,
\end{equation}
where $s(TQP)$ and $i(TQP)$ are the slope and intercept values at a constant TQP. To validate this relationship, PCs in the training set were selected and encoded with constant TQP values. Furthermore, Fig.~\ref{fig:TC_TBPP} depicts the linear relationship between TBPP and TC for a range of constant TQP values: 28, 34, 40, 46 and 51. And Table~\ref{tab:TC_TBPP_PLCC} presents the PLCC values between TBPP and TC. $s(TQP)$ and $i(TQP)$ can be modeled by
\begin{equation}
    s(TQP)=a_1\cdot TQP^2+a_2\cdot TQP+a_3
\label{equ:slope_TQP}
\end{equation}
and
\begin{equation}
    i(TQP)=b_1\cdot TQP+b_2\ .
\label{equ:intercept_TQP}
\end{equation}
Fig.~\ref{fig:TCTBPP_Fit_TQP} shows the relationship between them. In equations (\ref{equ:slope_TQP}) and (\ref{equ:intercept_TQP}), $a_1$, $a_2$, $a_3$, $b_1$ and $b_2$ are constants obtained through a training by the least squares method. Consequently, we get
\begin{equation}
\begin{split}
    TC = & (a_1 \cdot TQP^2 + a_2 \cdot TQP + a_3) \cdot TBPP \\
    & + (b_1 \cdot TQP + b_2)\ .
\end{split}
\label{equ:TC_TQP_TBPP}
\end{equation}

% alpha_TC_tNSL3
\begin{figure}[t]
\centering
\includegraphics[width=0.76\linewidth]{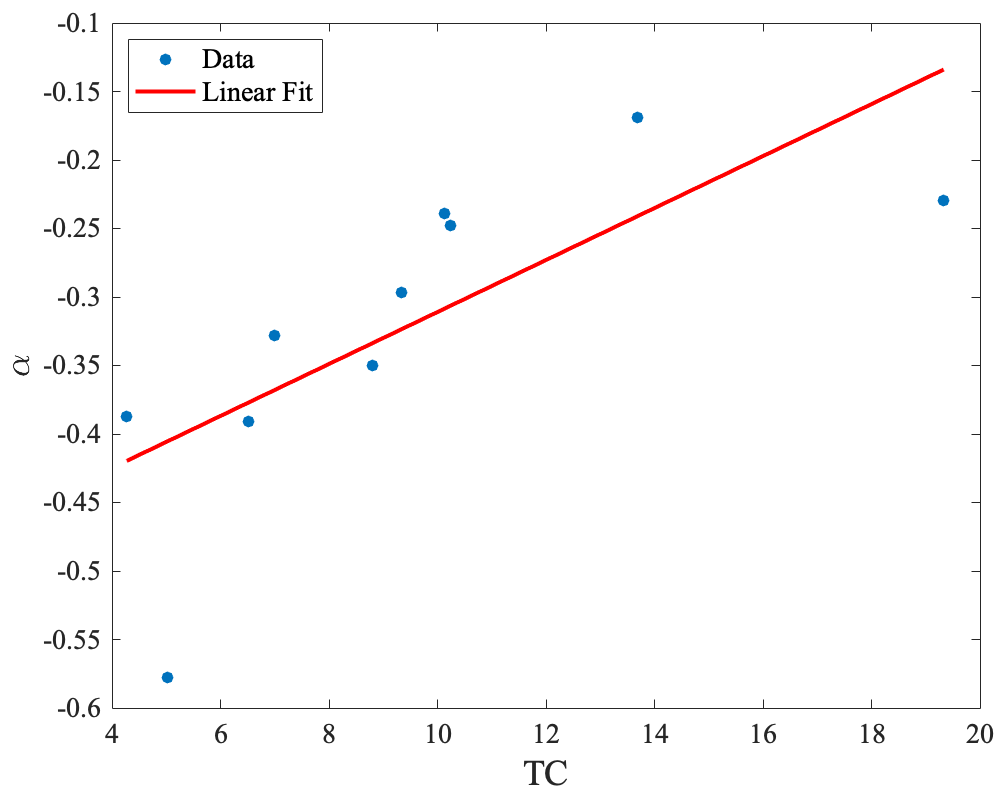}
\caption{Scatter plot of $\alpha$ and TC. The PLCC is 0.8224.}
\label{fig:alpha_TC_tNSL3}
\end{figure}

% Parameter values
\begin{table*}[t] 
\renewcommand{\arraystretch}{1.3}
\centering
\caption{Parameter values of the proposed model.} 
\label{tab:Parameter_values}
\vspace{-6pt}
\scalebox{1}{
\begin{tabular}{c c c c c c c c c c c}  
\\
\toprule
\hline
\(b\) & \(\alpha\) & \(\beta\) & \(a_1\) & \(a_2\) &\(a_3\) &\(b_1\) &\(b_2\) &\(l_1\) & \(l_2\) &\(l_3\) \\  
\hline  
      90.3036  & 0.0189 & -0.5006 & 0.2442 & -15.3958 & 247.4869 & 0.1311 & -4.1114 & 19.2911 & -8.8925 & -18.1897 \\  
\hline
\bottomrule
\end{tabular}  
}
\end{table*}

\subsection{Texture Distortion Assessment}
As demonstrated in Fig.~\ref{fig:MOS_TQP_10}, there is a linear relationship between the MOS and TQP for each PC when subjected to the condition of lossless geometry compression:
\begin{equation}
    MOS_T = a \cdot TQP + b\ ,
\label{equ:MOST_TQP}
\end{equation}
where $a$ and $b$ are parameters derived empirically. In addition, $MOS_T$ signifies MOS of the PC that is attributable solely to texture distortion. Considering $b$ is the maximum of $MOS_T$ common to each PC, it is established as a constant specified in Table~\ref{tab:Parameter_values}. $a$ is dependent on content of PC. Fig.~\ref{fig:alpha_TC_tNSL3} presents a scatter plot illustrating the relationship between $a$ and TC in the training set. Generally, a higher value of $a$, for a PC, exhibits higher TC (e.g.,~\textit{glasses\_case}). Conversely, PCs with lower TC such as~\textit{tool\_box} exhibit correspondingly lower values of $a$. Consequently, we introduce an empirical function that allows $a$ to adjust according to the PC content. The empirical function is given by the following equation:
\begin{equation}
    a = \alpha \cdot TC + \beta\ ,
\label{equ:a_TC}
\end{equation}
where $\alpha$ and $\beta$ are obtained through a training process that employs least squares fitting in the training set. Ultimately, incorporating (\ref{equ:MOST_TQP}) and (\ref{equ:a_TC}), we obtain
\begin{equation}
    MOS_T = (\alpha \cdot TC + \beta) \cdot TQP + b\ .
\end{equation}
The model can be employed to estimate perceived quality of PC without geometry distortion and is vital to the development of PCQA models that account for both geometry and texture distortions.

\subsection{Influence of Geometry Distortion and streamPCQ-TL Model}
\begin{figure*}[htpb]
\centering
\includegraphics[width=0.9\linewidth]{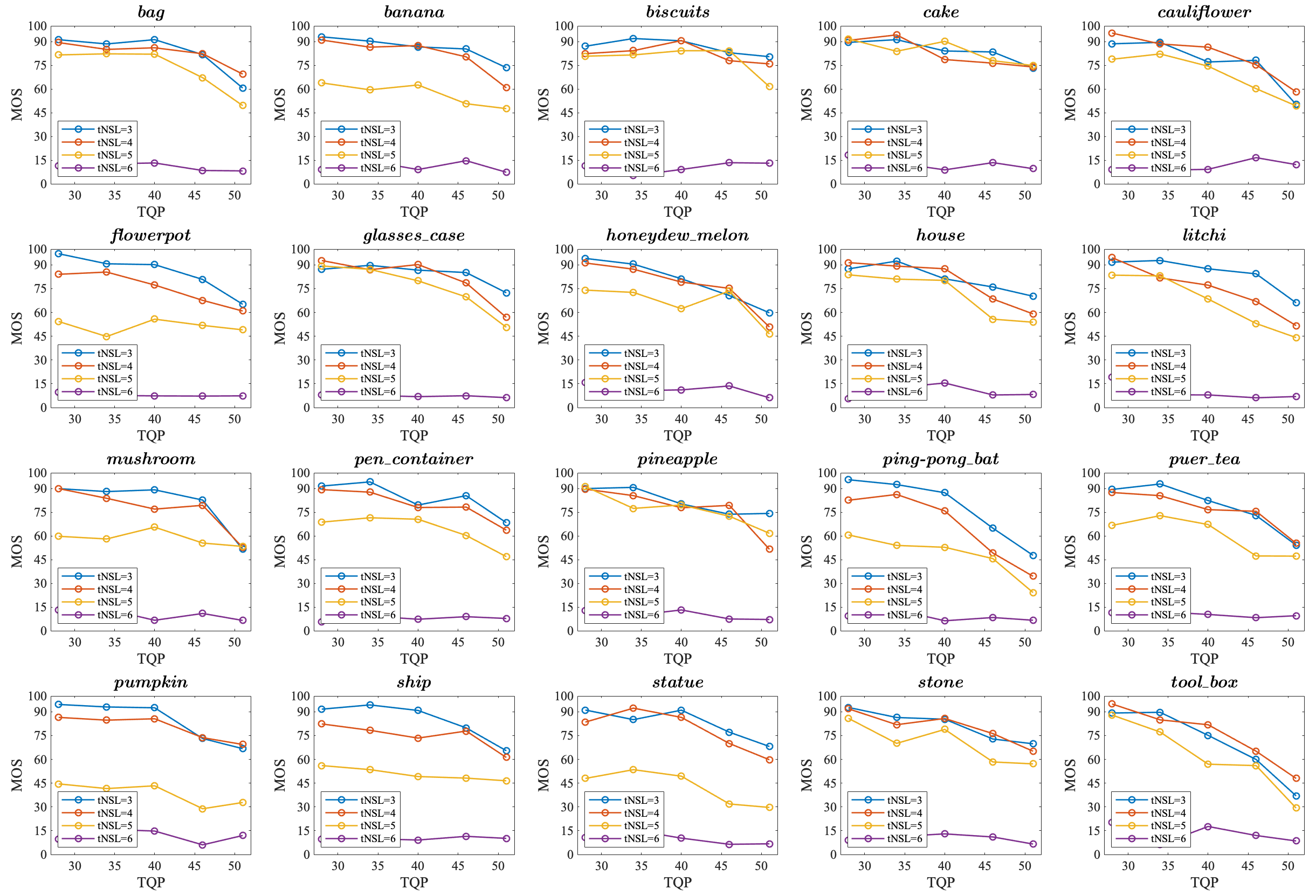}
\caption{MOS vs. TQP for various tNSLs.}
\label{fig:MOS_TQP_4x5}
\end{figure*}

\begin{figure*}[htpb]
\centering
\includegraphics[width=0.9\linewidth]{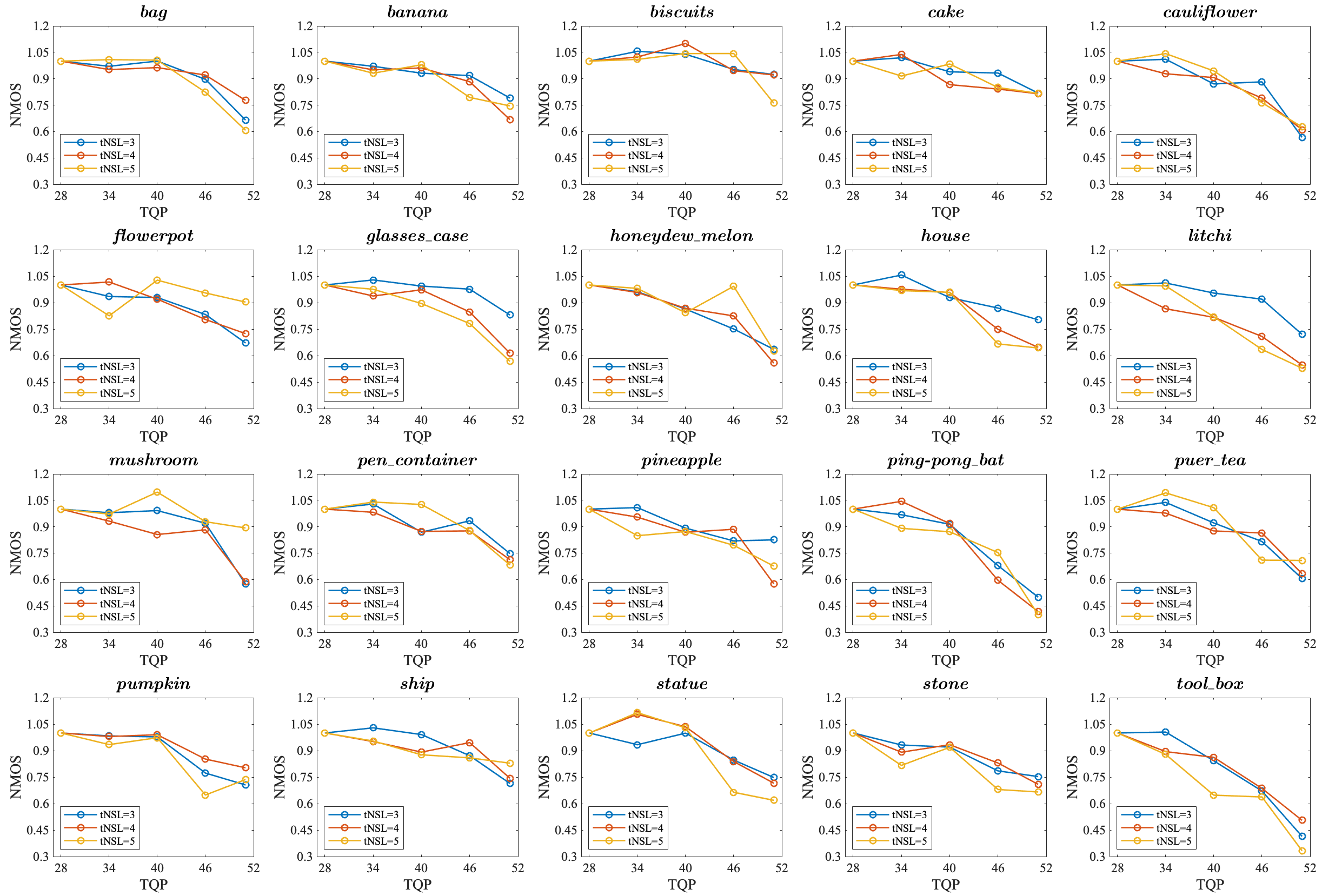}
\caption{Normalized MOS vs. TQP for various tNSLs.}
\label{fig:NMOS_TQP_4x5}
\end{figure*}

In the context of Trisoup encoding, a geometry coding technique that models object surfaces using a series of triangular meshes, the parameter tNSL plays a pivotal role in the process. This parameter dictates the size of the individual triangular nodes, thereby directly influencing the precision of the geometry representation. Fig.~\ref{fig:MOS_TQP_4x5} delineates the relationship between MOS and TQP at varying levels of tNSL for diverse PC contents. It is evident that, no matter the contents, MOS typically decreases as TQP increases at a fixed tNSL level and it diminishes with increments in tNSL at a constant TQP level. Now the critical question is whether the impact of TQP and tNSL is separable or not. To investigate the independence between the two factors, we introduce a normalized MOS (NMOS), defined as follows:
\begin{equation}
    NMOS(tNSL,TQP)=\frac{MOS(tNSL,TQP)}{MOS(tNSL,TQP_{min})}\ ,
\end{equation}
where $TQP_{min}$ denotes the minimum value of TQP in the experiment. Fig.~\ref{fig:NMOS_TQP_4x5} illustrates the relationship between NMOS and TQP across various levels of tNSL. Considering the MOSs exhibit a tendency to converge towards a constant value at tNSL=6 as illustrated in Fig.~\ref{fig:MOS_TQP_4x5} and the masking effect of HVS in this context predominantly subjected to the geometry distortion, we do not depict the curve at tNSL=6 representing the maximal geometry distortion in Fig.~\ref{fig:NMOS_TQP_4x5}. Apparently, the curves in Fig.~\ref{fig:NMOS_TQP_4x5} possess substantial overlaps in comparison with those depicted in Fig.~\ref{fig:MOS_TQP_4x5}, revealing appreciably independent influence of TQP and tNSL for MOS. Consequently, the impact of texture and geometry distortions can be largely regarded as separable factors. We employ the subsequent logistic function to model the geometry distortion as a attenuation factor of tNSL parameter:
\begin{equation}
    D_G(tNSL)=\frac{l_1}{1+e^{tNSL+l_2}}+l_3\ ,
\label{equ:Dg_tNSL}
\end{equation}
where $l_1$, $l_2$ and $l_3$ are empirically determined parameters and $D_G(tNSL)$ denotes the attenuation level. Based on the impact of texture distortion and geometry distortion on MOS is roughly dissociable, we introduce the streamPCQ-TL model: 
\begin{equation}
\begin{split}
    MOS_{est}= & MOS_T\cdot D_G=[(\alpha\cdot TC+\beta)\cdot TQP+b]\\
    & \cdot(\frac{l_1}{1+e^{tNSL+l_2}}+l_3)\ ,
\end{split}
\end{equation}
where TC is estimated by TQP and TBPP according to equation (\ref{equ:TC_TQP_TBPP}).

% MOS_TQP_4x5

% NMOS_TQP_4x5

\section{Experiment Results and Discussion}
\label{sec:Experimental Results and Discussion}

\subsection{Parameter Values of the Proposed Model}
The validation of the streamPCQ-TL model is conducted on the testing set including 10 PCs, i.e.,~\textit{banana},~\textit{biscuits},~\textit{cake},~\textit{flowerpot},~\textit{litchi},~\textit{pen\_container},~\textit{pumpkin},~\textit{ship},~\textit{statue} and~\textit{stone} which are rich both in geometry and texture features. It should be noted that these PCs have no overlap with those in the training set. Table~\ref{tab:Parameter_values} presents the model parameters derived from a least squares fitting on the training set. With specificity, $b$ is the average of MOSs when texture and geometry distortions are minimal. The values for $a_1$, $a_2$, $a_3$, $b_1$ and $b_2$ are obtained by equation (\ref{equ:slope_TQP}) and (\ref{equ:intercept_TQP}). Parameters $\alpha$ and $\beta$ are determined by equation (\ref{equ:a_TC}) while $l_1$, $l_2$ and $l_3$ are derived according to equation (\ref{equ:Dg_tNSL}). Once determined, these parameters are established as constants throughout the subsequent experiment results presented in this paper. For PCs originating from other databases, it may be necessary to adjust these parameter values.

\subsection{Leave-one-out Cross-validation}
On the subject of preventing overfitting and assessing the generalization performance of the model, we employ an approach, content-sensitive leave-one-out cross-validation (LOOCV), to derive the overall performance of the streamPCQ-TL model. In the LOOCV process, with $k$ set to 20 in this study, $k$ is defined the number of reference PCs. For each iteration, $k-1$ instances are utilized to train a model while the remaining instance serves as the test case. Each test case comprises 20 distorted PCs corresponding to their reference PC. Subsequently, the training and testing process is iterated k times, employing distinct partitions of the PCs for each time. To evaluate the performance, three classical statistical criterion are applied containing PLCC, SRCC and the root mean square error (RMSE) which measures the correlation and deviation between the objective scores generated by streamPCQ-TL model and the MOSs. Concurrently, a PLCC or SRCC value approaching 1 and an RMSE value approaching 0 signify a strong correlation with the subjective assessments, thereby denoting a higher degree of model accuracy. It is crucial to recognize that we compute values of the these statistical measures based on the test case and averaged the results to characterize the general performance of the proposed model. With the results of each cases presented in Table~\ref{tab:LOOCV}, we can observed that the performance in each test case even in the mean and standard deviation is great, a persuasive piece of evidence that the proposed model own satisfying performance in generalization and robustness.

% LOOCV_Results
\begin{table}[htpb]
\renewcommand{\arraystretch}{1.3}
\centering
\caption{LOOCV results on the proposed WPC6.0 database.} 
\label{tab:LOOCV}
\begin{tabular}{c | c c c }  
\toprule
\hline
Name & PLCC & SRCC & RMSE\\
\hline  
\emph{bag}              & 0.9571 & 0.8541 & 8.9721\\
\emph{banana}           & 0.9884 & 0.9444 & 4.4962\\
\emph{biscuits}         & 0.9592 & 0.7805 & 8.4273\\
\emph{cake}             & 0.9459 & 0.6842 & 9.7856\\
\emph{cauliflower}      & 0.9481 & 0.7850 & 9.5729\\
\emph{flowerpot}        & 0.9931 & 0.9594 & 3.4891\\
\emph{glassess\_case}   & 0.9533 & 0.7789 & 9.1967\\
\emph{honeydew\_melon}  & 0.9537 & 0.8391 & 9.0418\\
\emph{house}            & 0.9623 & 0.8105 & 8.2731\\
\emph{litchi}           & 0.9587 & 0.8767 & 8.5217\\
\emph{mushroom}         & 0.9686 & 0.8797 & 7.5040\\
\emph{pen\_container}   & 0.9778 & 0.8977 & 6.3251\\
\emph{pineapple}        & 0.9449 & 0.7774 & 9.9004\\
\emph{ping-pong\_bat}   & 0.9525 & 0.9038 & 9.1463\\
\emph{puer\_tea}        & 0.9665 & 0.8962 & 7.8065\\
\emph{pumpkin}          & 0.9816 & 0.9398 & 5.7518\\
\emph{ship}             & 0.9952 & 0.9549 & 2.9183\\
\emph{statue}           & 0.9786 & 0.9263 & 6.1774\\
\emph{stone}            & 0.9650 & 0.8511 & 8.0129\\
\emph{tool\_box}        & 0.9539 & 0.7880 & 9.1079\\
\hline
Mean                    & 0.9652 & 0.8564 & 7.6214\\
Standard deviation      & 0.0156 & 0.0748 & 2.0882\\
\hline
\bottomrule
\end{tabular}
\end{table}

\subsection{Ablation Test}
With ablation tests conducted, we investigate the contributions of the texture distortion and geometry distortion in the streamPCQ-TL model. Within the process of ablation tests, the training and testing procedures are same as those in streamPCQ-TL, which are employed to establish the texture-distortion-solely $MOS_T(TQP,\ TBPP)$ model and geometry-distortion-solely geometry attenuation factor $D_G(tNSL)$. As illustrated in Table~\ref{tab:ablation}, the tests results suggest that TQP, TBPP and tNSL play an critical role and make important contributions in the streamPCQ-TL model.

% Table: ablation
\begin{table}[t]  
\renewcommand{\arraystretch}{1.3}
\centering
\caption{Ablation test results on the proposed WPC6.0 database.} 
\label{tab:ablation}
\begin{tabular}{c | c c c }  
\toprule
\hline
PCQA model & PLCC & SRCC & RMSE\\
\hline  
% $MOS_T (TQP)$             & 0.6070 & 0.5726 & 17.4850\\
$MOS_T (TQP,\ TBPP)$    & 0.4030 & 0.3139 & 30.2074\\
$D_G(tNSL)$              & 0.9221 & 0.7806 & 11.9968\\
$\operatorname{\textbf{streamPCQ\text{-}TL(ours)}}$          & \textbf{0.9562} & \textbf{0.8589} & \textbf{8.3304}\\
\hline
\bottomrule
\end{tabular}  
\end{table}

% Pic:Random1000_PLCC_SRCC_RMSE
\begin{figure}[t]
\centering
\subfloat[]
{\includegraphics[width=0.49\linewidth]
{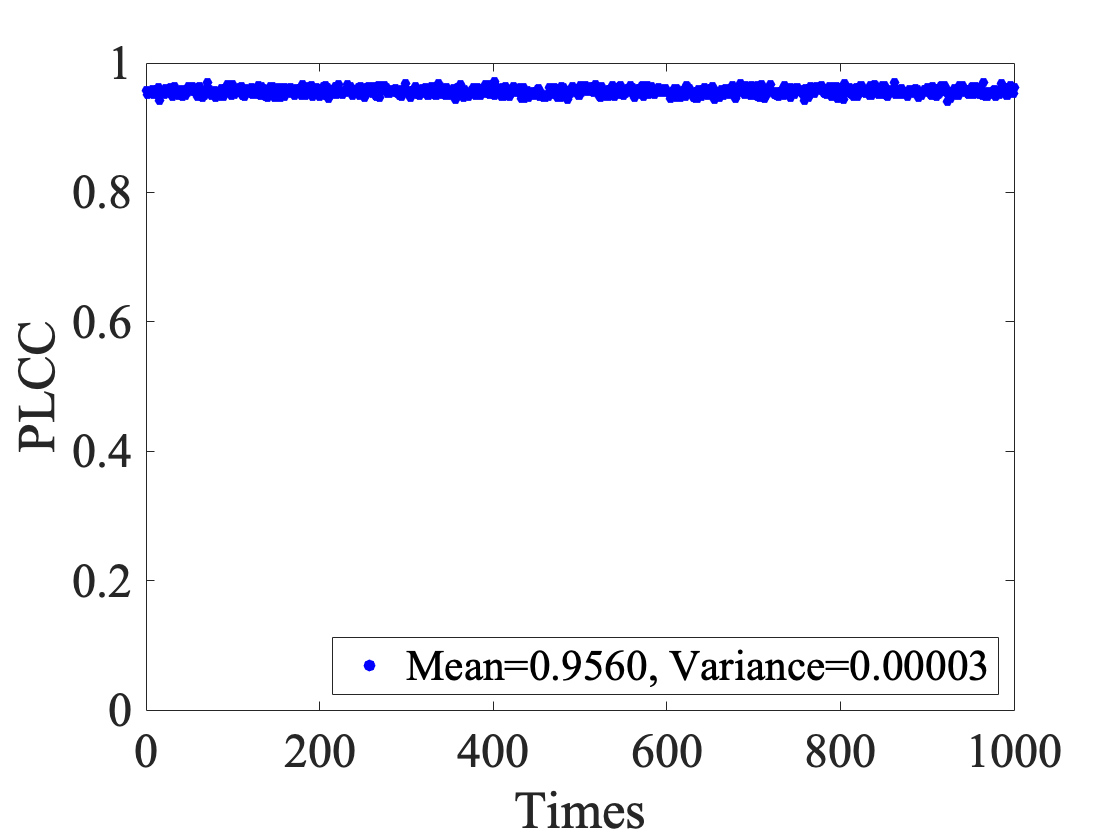}}\hskip0.01em
% \\
\subfloat[]
{\includegraphics[width=0.49\linewidth]
{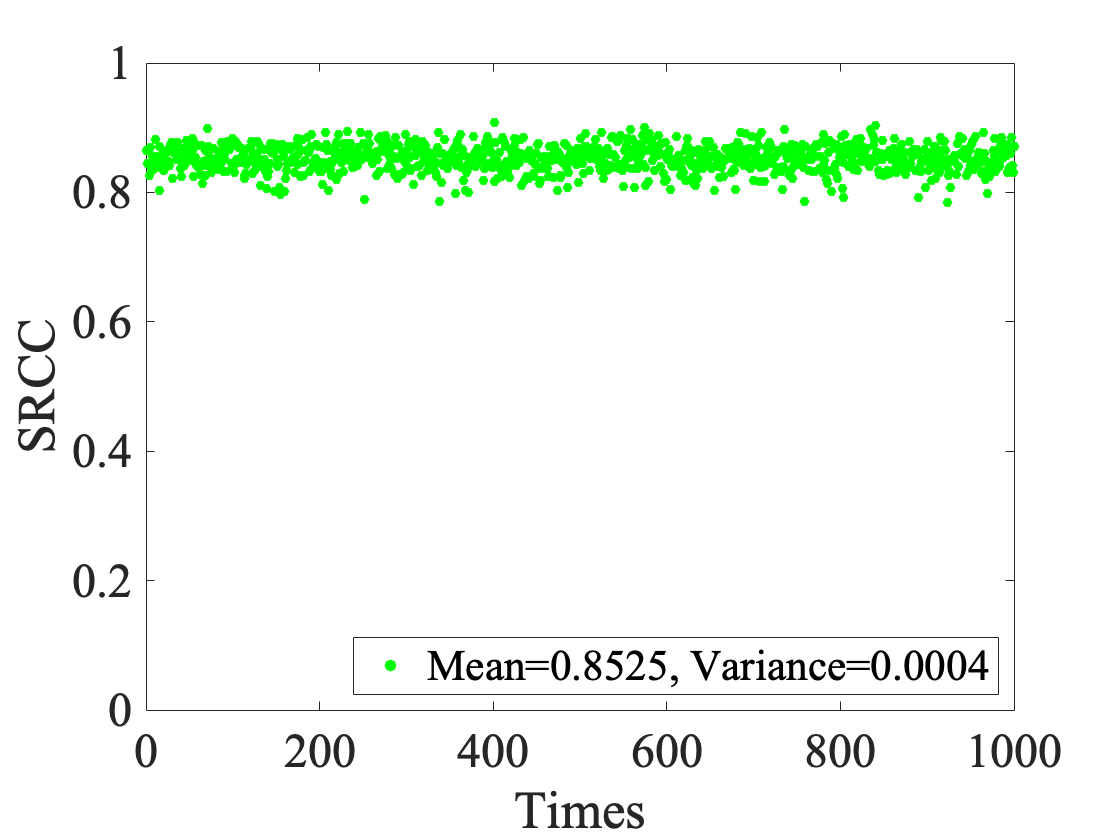}}\hskip0.01em
% \\
\subfloat[]
{\includegraphics[width=0.49\linewidth]{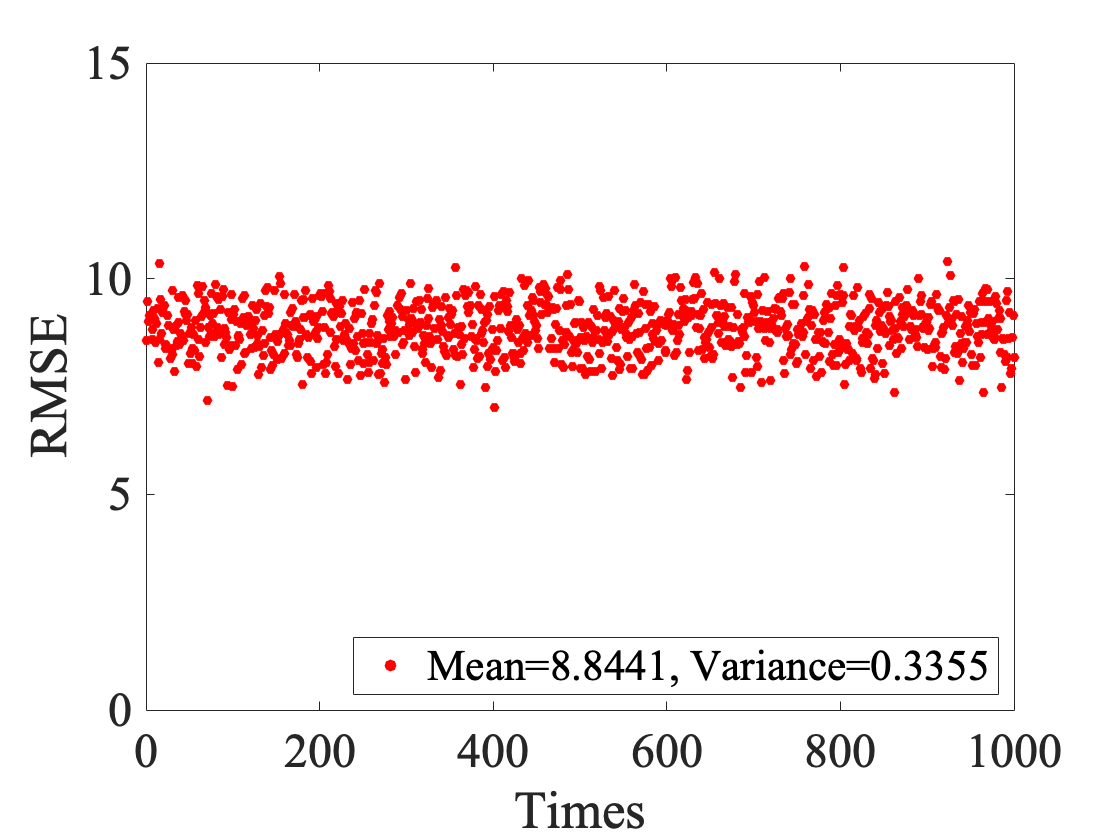}}\hskip0.01em
\caption{1000 randomized trials PLCC, SRCC and RMSE results on the proposed WPC6.0 database.}
\label{fig:random1000_PSR}
\end{figure}

\subsection{Reliability of the Proposed Method}
To substantiate the stability of the streamPCQ-TL model, we conduct an analysis on WPC6.0 database which is executed 1000 iterations with randomly partitioned training and testing sets, calculating the PLCC, SRCC and RMSE between the MOSs and objective scores. Specifically, we divide all distorted PCs into 20 parts based on their corresponding reference PCs. There are 20 distorted PCs for each reference PC. In the random experiment, we randomly select 10 parts for training and the remaining 10 parts for testing. Therefore, the component of PCs for training and testing are distinct for each iteration. The results displayed in Fig.~\ref{fig:random1000_PSR} demonstrate the durability of the established method.

% PCQA metrics Comparison on WPC6.0, M-PCCD and ICIP2020
\begin{table*}[t]
    \renewcommand{\arraystretch}{1.3}
    \centering
    \caption{Performance comparison of streamPCQ-TL against existing models. Results are formatted as follows: the highest rank is in \textbf{bold}, the second is \ul{underlined} and the third is with a \uwave{wavy underline}.}
    \label{tab:PCQA_Metrics_Performance_On_3_Dataset}
    \scalebox{1}{
        \begin{tabular}{c|c|c c c|c c c|c c c}
        \toprule
        % \toprule
        % \midrule
        \hline
        \multirow{2}{*}{Type} &\multirow{2}{*}{Metrics} &\multicolumn{3}{c|}{WPC6.0 (proposed)} &\multicolumn{3}{c|}{M-PCCD~\cite{alexiou2019comprehensive}} &\multicolumn{3}{c}{ICIP2020~\cite{perry2020quality}}\\
        \cline{3-11}
        & &PLCC &SRCC &RMSE &PLCC &SRCC &RMSE &PLCC &SRCC &RMSE\\ 
        \hline
        FR &$\operatorname{GraphSIM}$~\cite{yang2020inferring}  &0.8657 &0.8637 &15.5252 &\ul{0.9442} &\textbf{0.9429} &\ul{0.4099} &0.8585 &0.8617 &0.5703\\
           &$\operatorname{MS-GraphSIM}$~\cite{zhang2021ms}     &0.8847 &\uwave{0.8749} &14.4553 &0.9213 &0.8968 &0.5075 &0.8151 &0.7956 &0.6443\\
           &$\operatorname{MPED}$~\cite{yang2022mped}           &0.7684 &0.7199 &19.8492 &0.7939 &0.8148 &0.8368 &0.8498 &0.8314 &0.5862\\
           &$\operatorname{PointSSIM}_{\sigma^2}$~\cite{alexiou2020towards} &0.2957 &0.3626 &29.6264 &0.9135 &0.8965 &0.4559 &0.5865 &0.5575 &0.9008\\
           &$\operatorname{IW-SSIM}_p$~\cite{liu2022perceptual} &0.9167 &0.8480 &12.3893 &0.6956 &0.7533 &0.7228 &\ul{0.8919} &0.8608 &0.5029\\
           &$\operatorname{PSNR}_{p2po,M}$~\cite{tian2017evaluation}~\cite{tian2017updates} &0.9156 &0.7603 &12.4721 &0.8175 &0.8282 &0.6017 &0.8826 &\ul{0.8914} &0.5229\\
           &$\operatorname{PSNR}_{p2po,H}$~\cite{tian2017evaluation}~\cite{tian2017updates} &0.6025 &0.3706 &24.7531 &0.8700 &0.4354 &0.7911 &0.8388 &0.8305 &0.6055\\
           &$\operatorname{PSNR}_{Y}$~\cite{mekuria2016evaluation}~\cite{mekuria2016design} &0.4219 &0.4793 &28.1182 &0.7191 &0.6576 &0.8128 &0.7539 &0.7388 &0.7307\\
        \hline   
        RR &$\operatorname{PCM}_{RR}$~\cite{viola2020reduced}   &0.1271 &0.1563 &31.0138 &0.8323 &0.8251 &1.3223 &0.7586 &0.8463 &1.1122\\
        \hline
        NR &$\operatorname{3DTA}$~\cite{zhu20243dta}      &\textbf{0.9779} &\textbf{0.9389} &\textbf{6.4908} &0.8371 &0.8405 &0.7420 &0.7874 &0.7380 &0.6856\\
           &$\operatorname{GMS-3DQA}$~\cite{zhang2024gms} &0.9480 &0.8476 &9.8736 &\textbf{0.9715} &\ul{0.9339} &\textbf{0.3442} &\textbf{0.9981} &\textbf{0.9881} &\textbf{0.0686}\\
           &$\operatorname{MM-PCQA}$~\cite{zhang2022mm}   &\ul{0.9663} &\ul{0.9220} &\ul{7.9807} &0.6129 &0.6546 &1.0387 &0.7240 &0.5933 &0.7672\\
           &\textbf{streamPCQ-TL(ours)}  &\uwave{0.9562} &0.8589 &\uwave{8.3304} &\uwave{0.9308} &\uwave{0.9101} &\uwave{0.4168} &\uwave{0.8829} &\uwave{0.8908} &\ul{0.4329}\\
        \hline   
        \bottomrule   
        \end{tabular}
    }
\end{table*}

\subsection{Performance Comparison on the WPC6.0 Database}
In this part, we compare the streamPCQ-TL metric with 1) $\operatorname{GraphSIM}$~\cite{yang2020inferring}, 2) $\operatorname{MS-GraphSIM}$~\cite{zhang2021ms}, 3) $\operatorname{MPED}$~\cite{yang2022mped}, 4) $\operatorname{PointSSIM\ variance}$ ($\operatorname{PointSSIM}_{\sigma^2}$)~\cite{alexiou2020towards}, 5) $\operatorname{IW-SSIM}_p$~\cite{liu2022perceptual}, 6) $\operatorname{PCM}_{RR}$~\cite{viola2020reduced}, 7) Point-to-point mean squared error-based PSNR ($\operatorname{PSNR}_{p2po,M}$)~\cite{tian2017evaluation},~\cite{tian2017updates}, 8) Point-to-point Hausdorff distance-based PSNR ($\operatorname{PSNR}_{p2po,H}$)~\cite{tian2017evaluation},~\cite{tian2017updates}, 9) Point-to-point PSNR on color component ($\operatorname{PSNR}_{Y}$)~\cite{mekuria2016evaluation},~\cite{mekuria2016design}, 10) $\operatorname{3DTA}$~\cite{zhu20243dta}, 11) $\operatorname{GMS-3DQA}$~\cite{zhang2024gms} and 12) $\operatorname{MM-PCQA}$~\cite{zhang2022mm}, respectively. The results detailed in Table~\ref{tab:PCQA_Metrics_Performance_On_3_Dataset} observe that the proposed model maintain robust and competitive performance on the WPC6.0 database and similarly holds considerable performance even in NR metrics based on deep learning.

\subsection{Performance of streamPCQ-TL on Other Datasets}
To assess the generalizability of the streamPCQ-TL model, we conduct cross-database validation experiments on the M-PCCD~\cite{alexiou2019comprehensive} and the ICIP2020 database~\cite{perry2020quality}. Given the fact that Trisoup-Lifting encoded PCs in M-PCCD and ICIP2020 database are too deficient to train a model, we employ them exclusively to test the model trained on the WPC6.0 database, as well as against existing methods. With the results reported in Table~\ref{tab:PCQA_Metrics_Performance_On_3_Dataset}, it can be observed that the proposed model has competitive performance in comparison with advanced methods in the literature. It is worth noting that many PCs within the M-PCCD and ICIP2020 database are utilized for training the other models which are used in the comparison and such contents are completely unforeseen to streamPCQ-TL model during its establishment. Specifically, the M-PCCD database shares common PCs with those used to develop GraphSIM such as \textit{longdress}, \textit{loot} and \textit{romanoillmap}. Additionally, some PCs were used in the training phase of GMS-3DQA in the comparison experiment on M-PCCD, which were unbeknown to streamPCQ-TL model. Similarly, some PCs within ICIP2020 also were employed for training GMS-3DQA in corresponding experiment, which were not anticipated by the streamPCQ-TL model.  Furthermore, it is also worth mentioning that the NR streamPCQ-TL model working at bitstream-layer has low computational complexity detailed in Section~\ref{sec:bitstreamPCQA}.

% Salience_on_Three_Database
% [htpb]
\begin{figure}[t]
\centering
\subfloat[WPC6.0 (proposed)]
{\includegraphics[width=0.49\linewidth]
{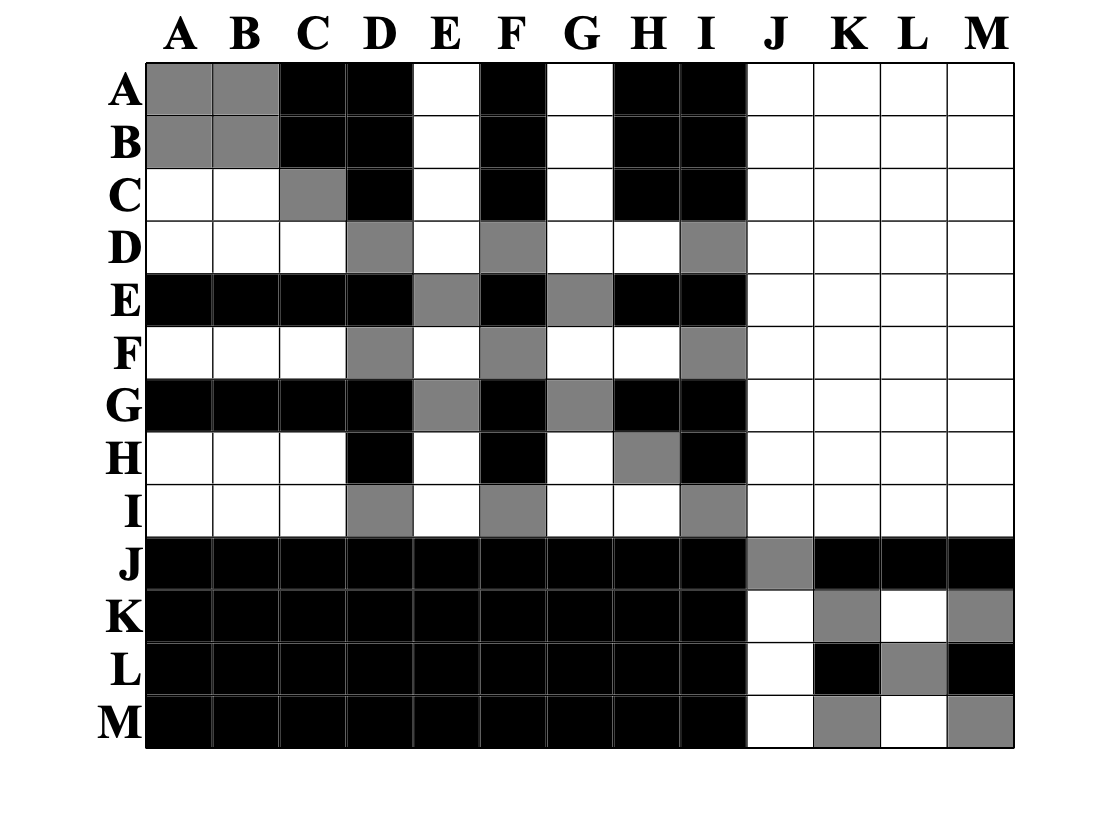}}\hskip0.01em
\subfloat[M-PCCD~\cite{alexiou2019comprehensive}]
{\includegraphics[width=0.49\linewidth]
{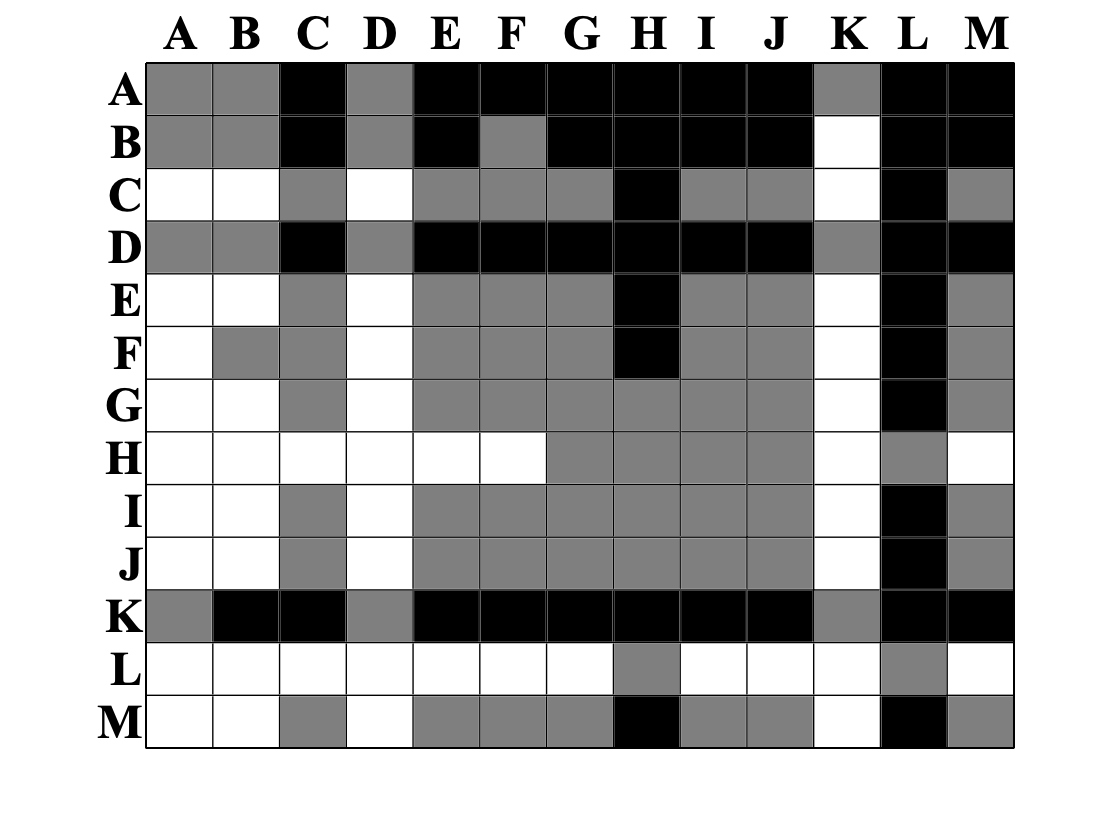}}\hskip0.01em
\subfloat[ICIP2020~\cite{perry2020quality}]
{\includegraphics[width=0.49\linewidth]
{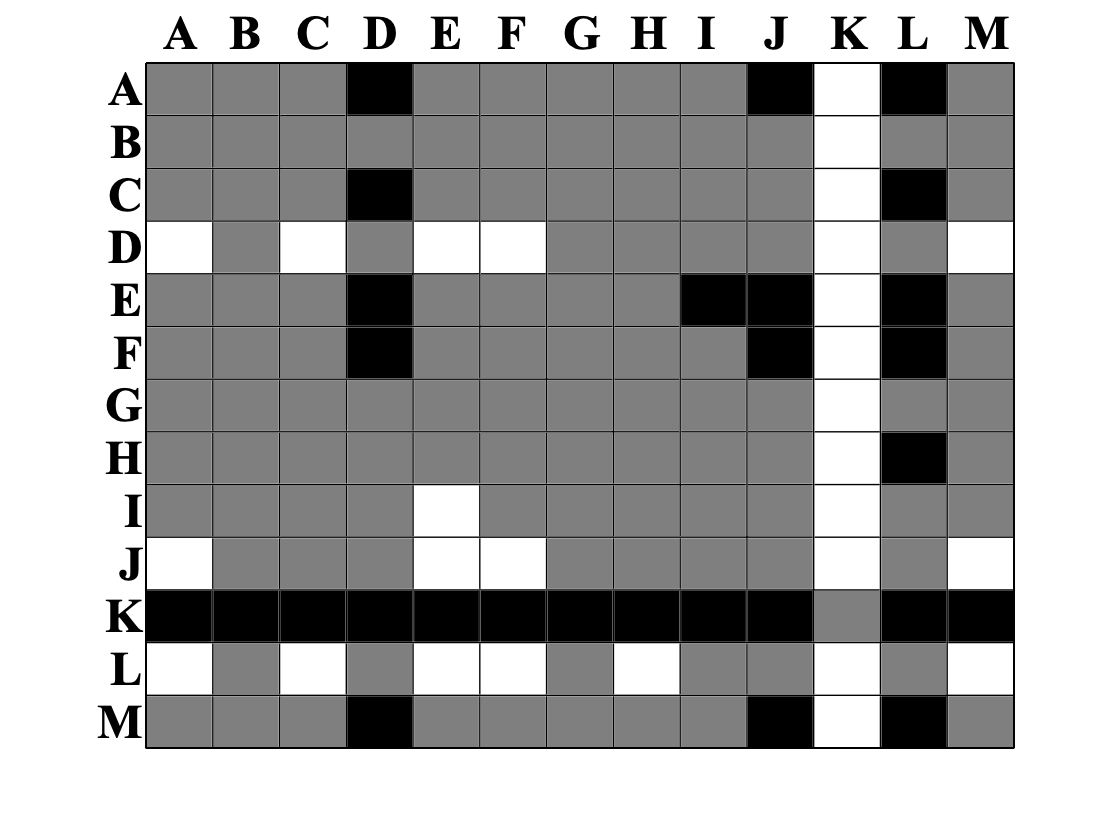}}\hskip0.01em

\caption{Statistical significance analysis on three PC datasets. A: $\operatorname{GraphSIM}$~\cite{yang2020inferring}; B: $\operatorname{MS-GraphSIM}$~\cite{zhang2021ms}; C: $\operatorname{MPED}$~\cite{yang2022mped}; D: $\operatorname{PointSSIM}_{\sigma^2}$~\cite{alexiou2020towards}; E: $\operatorname {IW-SSIM}_p$~\cite{liu2022perceptual}; F: $\operatorname{PCM}_{RR}$~\cite{viola2020reduced}; G: $\operatorname{PSNR}_{p2po,M}$~\cite{tian2017evaluation},~\cite{tian2017updates}; H: $\operatorname{PSNR}_{p2po,H}$~\cite{tian2017evaluation},~\cite{tian2017updates}; I: $\operatorname{PSNR}_{Y}$~\cite{mekuria2016evaluation},~\cite{mekuria2016design}; J: $\operatorname{3DTA}$~\cite{zhu20243dta}; K: $\operatorname{GMS-3DQA}$~\cite{zhang2024gms}; L: $\operatorname{MM-PCQA}$~\cite{zhang2022mm}; \textbf{M}: \textbf{streamPCQ-TL(ours)}.} 
\label{fig:Salience_on_Three_Database}
\end{figure}

\subsection{Statistical Significance Analysis}
To assure the proposed model is statistically significant, a rigorous statistical significance analysis on WPC6.0, M-PCCD~\cite{alexiou2019comprehensive} and ICIP2020~\cite{perry2020quality} was conducted by employing a method introduced in~\cite{sheikh2006statistical}. During this process, a nonlinear regression function used, the objective quality scores are mapped onto the subjective scores. On the subject of the prediction residuals all have zero-mean, a model with lower variance is normally thought superior to one with higher variance, as it indicates a more precise fit to the data. Considering hypothesis testing is conducted by making use of F-statistics, provided the sample size is sufficiently large, specifically exceeding 30 in this experiment, we can draw upon the central limit theorem~\cite{montgomery2020applied} to presume that the residuals are normally distributed. The test statistic, defined as the ratio of variances, is utilized under the null hypothesis that the prediction residuals of one quality model are stem from an identical distribution to those of another model, asserting their statistical equivalence at a 95\% confidence level. Objective models are compared against each other, with the comparative results depicted in Fig.~\ref{fig:Salience_on_Three_Database}, where a black block indicates that the row model significantly outperforms the column model while a white block signifies the converse and a gray block denotes that the two models are statistically indistinguishable. More specifically, Fig.~\ref{fig:Salience_on_Three_Database}(a) illustrates the results for each metrics in the literature on the WPC6.0 database, Fig.~\ref{fig:Salience_on_Three_Database}(b) for M-PCCD and Fig.~\ref{fig:Salience_on_Three_Database}(c) for ICIP2020. In Fig.~\ref{fig:Salience_on_Three_Database}, the labels 'A' to 'L' denote different existing metrics specifically $\operatorname{GraphSIM}$~\cite{yang2020inferring}, $\operatorname{MS-GraphSIM}$~\cite{zhang2021ms}, $\operatorname{MPED}$~\cite{yang2022mped}, $\operatorname{PointSSIM}_{\sigma^2}$~\cite{alexiou2020towards}, $\operatorname {IW-SSIM}_p$~\cite{liu2022perceptual}, $\operatorname{PCM}_{RR}$~\cite{viola2020reduced}, $\operatorname{PSNR}_{p2po,M}$~\cite{tian2017evaluation},~\cite{tian2017updates}, $\operatorname{PSNR}_{p2po,H}$~\cite{tian2017evaluation},~\cite{tian2017updates}, $\operatorname{PSNR}_{Y}$~\cite{mekuria2016evaluation},~\cite{mekuria2016design}, $\operatorname{3DTA}$~\cite{zhu20243dta}, $\operatorname{GMS-3DQA}$~\cite{zhang2024gms} and $\operatorname{MM-PCQA}$~\cite{zhang2022mm}, respectively. And the label 'M' signifies the proposed streamPCQ-TL model. It can be observed that the streamPCQ-TL model outperforms the majority of existing metrics and has competitive performance in contrast to those models like classical $\operatorname{GraphSIM}$ or which are based on deep learning techniques.

\begin{table}[t]
\centering
\fontsize{7}{10.7}\selectfont 
\caption{Time cost comparison of PCQA models.}
\label{tab:runtime}
% \begin{tabular}{|p{1.25cm}|p{1.25cm} |p{1.25cm}| p{1.25cm} |p{1.4cm}|}
\begin{tabular}{
   >{\centering\arraybackslash}p{1.4cm}|
   >{\centering\arraybackslash}m{1.7cm} 
   >{\centering\arraybackslash}p{1.1cm} 
   >{\centering\arraybackslash}p{1.0cm} 
   >{\centering\arraybackslash}p{1.25cm} 
}
\toprule
\hline
Content \ (point number) & PCQA model & Execution time (Seconds) & Execution time ($10^{-7}$ seconds \ per point) & Relative time to \textbf{streamPCQ-TL} \\ 
\hline

\multirow{5}{=}{\centering \textit{ship}\\(684617)}
&GraphSIM & 145.95 & 2131.84 & 13268.12 \\
&MS-GraphSIM & 153.98 & 2249.11 & 13998.00 \\
&MPED & 140.59 & 2053.55 & 12780.86 \\
&PointSSIM$_{\sigma^2}$ & 29.30 & 427.96 & 2663.54 \\
&IW-SSIM$_p$ & 57.56 & 840.80 & 5233.00 \\
&PSNR$_Y$ & 4.22 & 61.60 & 383.36 \\
&PCM$_{RR}$ & 215.82 & 3152.38 & 19619.76 \\
&\textbf{streamPCQ-TL} & \textbf{0.011} & \textbf{0.161} & \textbf{1} \\
\hline

\multirow{5}{=}{\centering \textit{statue}\\(1637577)}  
&GraphSIM & 405.84 & 2478.28 & 22546.53 \\
&MS-GraphSIM & 464.22 & 2834.83 & 26790.27 \\
&MPED & 343.22 & 2095.92 & 19067.98 \\
&PointSSIM$_{\sigma^2}$ & 89.54 & 546.77 & 4974.35 \\
&IW-SSIM$_p$ & 135.77 & 829.06 & 7542.51 \\
&PSNR$_Y$ & 12.88 & 78.64 & 715.40 \\
&PCM$_{RR}$ & 659.74 & 4028.73 & 36651.98 \\
&\textbf{streamPCQ-TL} & \textbf{0.018} & \textbf{0.110} & \textbf{1} \\
\hline
\multirow{5}{=}{\centering \textit{pen\_container}\\(2878318)} 
&GraphSIM & 653.04 & 2268.82 & 28393.03 \\
&MS-GraphSIM & 685.59 & 2381.91 & 29808.25 \\
&MPED & 608.14 & 2112.84 & 26440.96 \\
&PointSSIM$_{\sigma^2}$ & 137.33 & 477.11 & 5970.76 \\
&IW-SSIM$_p$ & 200.20 & 695.55 & 8704.37 \\
&PSNR$_Y$ & 19.76 & 68.65 & 859.17 \\
&PCM$_{RR}$ & 1004.56 & 3490.09 & 43676.48 \\
&\textbf{streamPCQ-TL} & \textbf{0.023} & \textbf{0.080} & \textbf{1} \\
\hline

\bottomrule
\end{tabular}
\end{table}

\subsection{Time Complexity Analysis}
To validate the proposed streamPCQ-TL model has low time complexity, we compare its execution time on three PCs, which represent different scales on content complexity of PC, specifically low, medium and high levels, against with some metrics discussed in this paper. The device performed the analysis is a workstation equipped with an Intel(R) Xeon(R) Gold 6253CL CPU @3.10GHz, a 128GB SAMSUNG MZ7LH480HAHQ-00005 RAM, a ST4000NM000A-2HZ100 hard disk, a NVDIA GeForce RTX 3090 graphics card and a Windows 10 Professional operating system. The results are showed in Table~\ref{tab:runtime}.

It can be observed the higher content complexity a PC owns, the more time ours model saves. And the time each model takes, showed in Table~\ref{tab:runtime}, increases with content complexity of PC. Apparently, the proposed streamPCQ-TL model has mega and striking advantages on runtime and is the lowest one on time cost, compared with those metrics in Table~\ref{tab:runtime}, particularly much faster than $\operatorname{GraphSIM}$~\cite{yang2020inferring}, $\operatorname{MPED}$~\cite{yang2022mped} and $\operatorname{PCM}_{RR}$~\cite{viola2020reduced}. Therefore, the developed streamPCQ-TL model is more apposite for real-time or time-critical applications in practical scenarios due to its NR property and low time complexity.

\section{Conclusion}\label{sec:conclusion}
In this study, we have addressed the problem of PCQA dedicated to Trisoup-Lifting encoded PCs. Without fully decoding, parameters texture quantization parameter (TQP), texture bits per point (TBPP) and trisoupNodeSizeLog2 (tNSL) are extracted from the PC bitstream to establish the proposed bitstream-layer NR-PCQA model named streamPCQ-TL. Texture complexity estimation, texture distortion assessment and geometry distortion model are modeled respectively using the above key parameters before the overall PCQA model is reached. Compared to state-of-the-art FR, RR and NR PCQA metrics, the streamPCQ-TL model reveals highly robust and competitive performance at a fraction of computational cost, which makes it a more apposite choice in time-critical applications. Additionally, for the absence of a large-scale PCQA database exclusively encoded by Trisoup-Lifting mode and its corresponding labels more particularly MOSs derived from subjective perceptual quality assessment, we have established the WPC6.0 database, the first database dedicated to Trisoup-Lifting encoding mode, which is not only a platform for training and testing the developed streamPCQ-TL model but a choice to validate other PCQA models both currently and in the day to come. For further research, the source code and database will be publicly released.

\bibliographystyle{IEEEtran}
\bibliography{ref_JC}

\end{document}